\documentclass{article}

\PassOptionsToPackage{linesnumbered,ruled,vlined}{algorithm2e}
\PassOptionsToPackage{english}{babel}
\PassOptionsToPackage{capitalise,noabbrev}{cleveref}
\PassOptionsToPackage{pdftex}{graphicx}
\PassOptionsToPackage{bookmarks}{hyperref}
\PassOptionsToPackage{utf8}{inputenc}
\PassOptionsToPackage{newfloat}{minted}
\PassOptionsToPackage{numbers,compress}{natbib}
\PassOptionsToPackage{dvipsnames}{xcolor}

\usepackage{adjustbox}
\usepackage{amsfonts}
\usepackage{amsmath}
\usepackage{amssymb}
\usepackage{babel}
\usepackage{bm}
\usepackage{booktabs}
\usepackage{cancel}
\usepackage{caption}
\usepackage{enumitem}
\usepackage{etoolbox}
\usepackage{fancyvrb}
\usepackage{graphicx}
\usepackage{inputenc}
\usepackage{listings}
\usepackage{mathtools}
\usepackage{microtype}
\usepackage{natbib}
\usepackage{nicefrac}
\usepackage{subfigure}
\usepackage{svg}
\usepackage{tikz}
\usepackage{verbatim}
\usepackage{wrapfig}
\usepackage{xcolor}

\usepackage{hyperref}[colorlinks=true, linkcolor=blue]
\usepackage{cleveref}

\usetikzlibrary{positioning,arrows.meta,quotes,shapes,bayesnet}
\tikzset{>=latex}

\crefname{equation}{}{}
\Crefname{equation}{}{}

\newtoggle{icml} \togglefalse{icml}
\newtoggle{neurips} \togglefalse{neurips}
\newtoggle{aistats} \togglefalse{aistats}

\newcommand{\determ}[1]{\ensuremath{\left\lvert#1\right\rvert}}  

\newcommand{\ci}{\ensuremath{\mathrel{\perp\mspace{-10mu}\perp}}}  

\newcommand{\mbf}[1]{{\ensuremath{\boldsymbol{\mathbf{#1}}}}}  

\usepackage{empheq}
\usepackage[most]{tcolorbox}
\newtcbox{\mymath}[1][]{%
    nobeforeafter, math upper, tcbox raise base,
    enhanced, colframe=black!30!black,
    colback=white!30, boxrule=1pt,
    #1}
\newtcbox{\mywboxtext}{on line,colback=white,colframe=black,size=fbox,arc=3pt,boxrule=0.8pt}
\newcommand{\mywboxmath}[1]{\mywboxtext{$#1$}}

\definecolor{darkgreen}{rgb}{0.0, 0.2, 0.13}

\definecolor{cornellred}{rgb}{0.7, 0.11, 0.11}
\newcommand{\marc}[1]{{\color{purple} Marc: #1}}

\renewcommand{\marc}[1]{}

\usepackage[algo2e,ruled,vlined]{algorithm2e}
\usepackage{dsfont}
\DeclareMathOperator{\position}{\mathbf q}
\DeclareMathOperator{\momentum}{\mathbf p}
\DeclareMathOperator{\state}{\mathbf z}

\newtheorem{theorem}{Theorem}

\usepackage{placeins}

\usepackage[accepted]{icml2021}

\icmltitlerunning{A Practical Method for Constructing Equivariant Multilayer Perceptrons for Arbitrary Matrix Groups}

\begin{document}

\twocolumn[
\icmltitle{A Practical Method for Constructing Equivariant Multilayer Perceptrons for Arbitrary Matrix Groups}
\icmlsetsymbol{equal}{*}



\icmlcorrespondingauthor{Marc Finzi}{maf820@nyu.edu}

\begin{center}
    \begin{tabular}{cccc}
         \textbf{Marc Finzi} & \textbf{Max Welling} &  \textbf{Andrew Gordon Wilson} \\
         New York University     & University of Amsterdam         & New York University
    \end{tabular}
\end{center}

\icmlkeywords{Equivariance, Symmetries, Machine Learning, ICML}

\vskip 0.3in
]



\printAffiliationsAndNotice{}  

\begin{abstract}
Symmetries and equivariance are fundamental to the generalization of neural networks on domains such as images, graphs, and point clouds. Existing work has primarily focused on a small number of groups, such as the translation, rotation, and permutation groups. In this work we provide a completely general algorithm for solving for the equivariant layers of matrix groups. In addition to recovering solutions from other works as special cases, we construct multilayer perceptrons equivariant to multiple groups that have never been tackled before, including  $\mathrm{O}(1,3)$, $\mathrm{O}(5)$, $\mathrm{Sp}(n)$, and the Rubik's cube group. 
Our approach outperforms non-equivariant baselines, with applications to particle physics and dynamical systems. We release our software library to enable researchers to construct equivariant layers for arbitrary matrix groups.
\end{abstract}

\section{Introduction}
\label{sec:intro}

As machine learning has expanded to cover more areas, the kinds of structures and data types we must accommodate grows ever larger. While translation equivariance may have been sufficient for working with narrowly defined sequences and images, with the  expanding scope to sets, graphs, point clouds, meshes, hierarchies, tables, proteins, RF signals, games, PDEs, dynamical systems, and particle jets, we require new techniques to exploit the structure and symmetries in the data.

In this work we propose a general formulation for equivariant multilayer perceptrons (EMLP). Given a set of inputs and outputs which transform according to finite dimensional representations of a symmetry group, we characterize all linear layers that map from one space to the other, and provide a polynomial time algorithm for computing them. We release our \href{https://github.com/mfinzi/equivariant-MLP}{\underline{library}}, along with \href{https://emlp.readthedocs.io/en/latest/}{\underline{documentation}}, and \href{https://colab.research.google.com/github/mfinzi/equivariant-MLP/blob/master/docs/notebooks/colabs/all.ipynb}{\underline{examples}}.

\begin{figure}[t]
    \centering
	\includegraphics[width=0.48\textwidth]{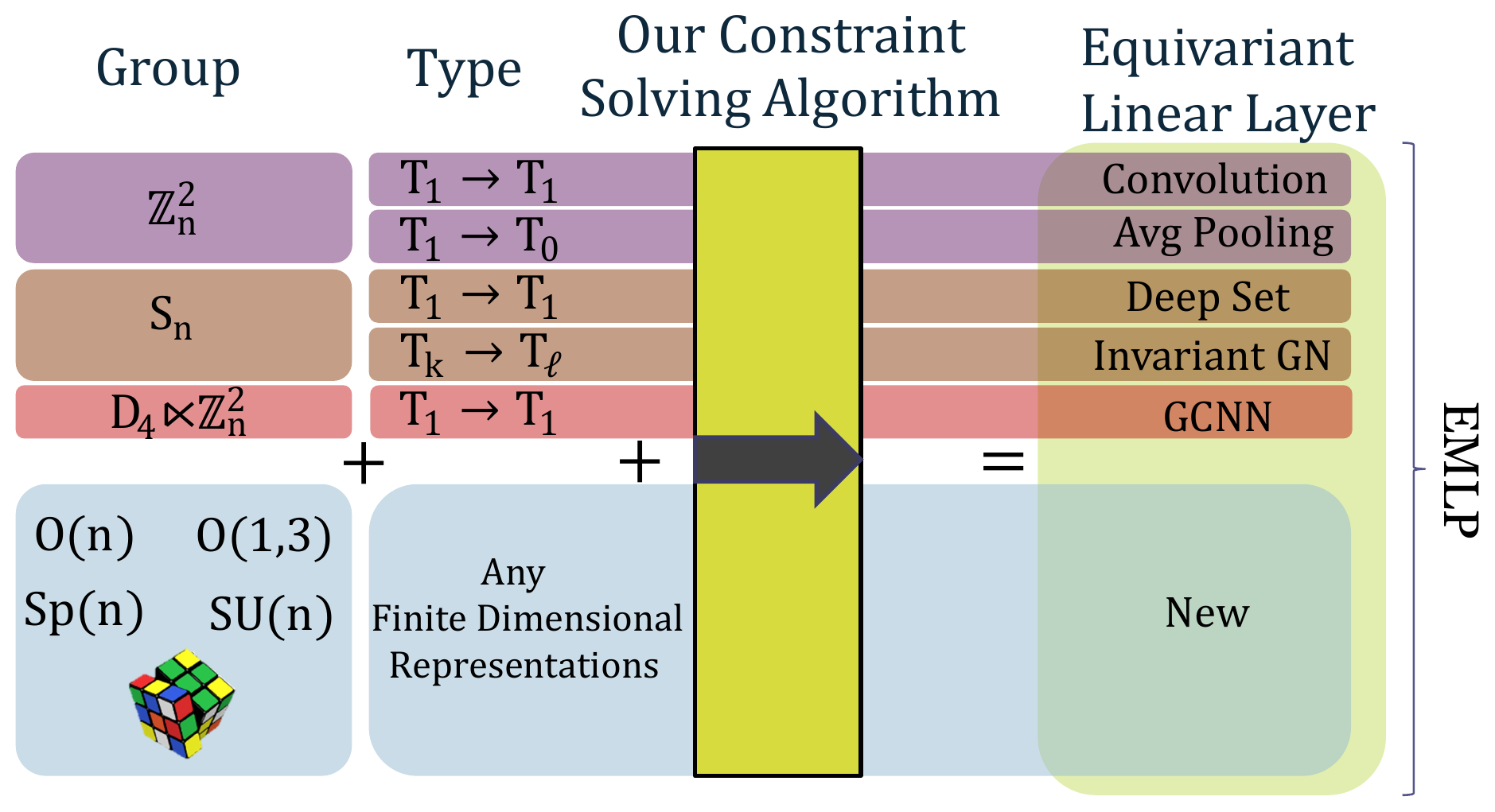}
	\caption{We provide a general and efficient method for solving equivariance constaints. For particular symmetry groups and type signatures, we recover other well known equivariant layers while also enabling application to new groups and representations. \marc{TODO: vectorize}}
    \label{fig:front_page}
\end{figure}

Figure~\ref{fig:front_page} illustrates how
the convolutional layers of a CNN \citep{lecun1989backpropagation}, the permutation equivariant deep sets \citep{zaheer2017deep}, graph layers \citep{maron2018invariant}, and layers for networks equivariant to point clouds \citep{thomas2018tensor}, all arise as special cases of our more general algorithm.

We summarize our contributions as follows:
\begin{itemize}
\item We prove that the conditions for equivariance to matrix groups with arbitrary linear representations can be reduced to a set of $M+D$ constraints, where $M$ is the number of discrete generators and $D$ is the dimension of the group. 

\item We provide a polynomial time algorithm for solving these constraints for finite dimensional representations, and we show that the approach can be accelerated by exploiting structure and recasting it as an optimization problem.

\item With the addition of a bilinear layer, we develop the Equivariant MultiLayer Perceptron (EMLP), a general equivariant architecture that can be applied to a new group by specifying the group generators.

\item Demonstrating the generality of our approach, we apply our network to multiple groups that were previously infeasible, such as the orthogonal group in five dimensions $\mathrm{O}(5)$, the full Lorentz group $\mathrm{O}(1,3)$, the symplectic group $\mathrm{Sp}(n)$, the Rubik's cube group, \emph{with the same underlying architecture}, outperforming non equivariant baselines.

\end{itemize}

\section{Related Work}
While translation equivariance in convolutional neural networks \citep{lecun1989backpropagation} has been around for many years, more general group equivariant neural networks were introduced in \citet{cohen2016group} for discrete groups with GCNNs. There have been a number of important works generalizing the approach to make use of the irreducible group representations for the continuous rotation groups
$\mathrm{SO}(2)$ \citep{cohen2016steerable,esteves2017polar,marcos2017rotation}, $\mathrm{O}(2)$ \citep{weiler2019general},
$\mathrm{SO}(3)$ \citep{thomas2018tensor,weiler20183d,anderson2019cormorant}, $\mathrm{O}(3)$ \citep{e3nn_tutorial_2020_3724982} and their discrete subgroups.The requirements and complexity of working with irreducible representations has limited the scope of these methods, with only one example outside of these two rotation groups with the identity component of the Lorentz group $\mathrm{SO}^+(1,3)$ in \citet{bogatskiy2020lorentz}.

Others have used alternate approaches for equivariance through group FFTs \citep{cohen2018spherical}, and regular group convolution \citep{worrall2019deep,bekkers2019b,finzi2020generalizing}. These methods enable greater flexibility; however, achieving equivariance for continuous groups with the regular representation is fundamentally challenging, since the regular representation is infinite dimensional.

Meanwhile, the theoretical understanding and practical methods for equivariance to the permutation group $S_n$ have advanced considerably for the application to sets \citep{zaheer2017deep}, graphs \citep{maron2018invariant}, and related objects \citep{serviansky2020set2graph}. Particular instances of equivariant networks have been shown to be \emph{universal}: with a sufficient size these networks can approximate equivariant functions for the given group with arbitrary accuracy \citep{maron2019universality,ravanbakhsh2020universal,dym2020universality}.

Despite these developments, there is still no algorithm for constructing equivariant networks that is completely general to the choice of symmetry group or representation. Furthest in this direction are the works of \citet{lang2020wigner}, \citet{ravanbakhsh2017equivariance}, and \citet{van2020mdp} with some of these ideas also appearing in \citet{WOOD199633}. Based on the Wigner-Eckert theorem, \citet{lang2020wigner} show a general process by which equivariant convolution kernels can be derived for arbitrary compact groups. However this process still requires considerable mathematical legwork to carry out for a given group, 
and is not applicable beyond compact groups.
\citet{ravanbakhsh2017equivariance} show how equivariance can be achieved by sharing weights over the orbits of the group, but is limited to regular representations of finite groups. Unlike \citet{lang2020wigner} and \citet{ravanbakhsh2017equivariance}, \citet{van2020mdp} present an explicit algorithm for computing equivariant layers. However, the complexity of this approach scales with the size of the group and quickly becomes too costly for large groups and impossible for continuous groups like $\mathrm{SO}(n),\mathrm{O}(1,3)$, $\mathrm{Sp}(n)$, and $\mathrm{SU}(n)$.

\section{Background}\label{sec:background}

In order to present our main results, we first review some necessary background on group theory. Most importantly, symmetry groups can be broken down in terms of discrete and continuous generators, and these can act on objects through group and Lie algebra representations.

\textbf{Finite Groups and Discrete Generators.}
A group $G$ is finitely generated if we can write each element $g\in G$ as a sequence from a discrete set of generators $\{h_1,h_2,...h_M\}$ and their inverses $h_{-k}=h_k^{-1}$. For example we may have an element $g=h_1h_2h_2h_1^{-1}h_3$ and can be written more compactly $g = \Pi_{i=1}^Nh_{k_i}$ for the integer sequence $k=[1,2,2,-1,3]$.

All finite groups, like the cyclic group $\mathbb{Z}_n$, the dihedral group $\mathrm{D}_n$, the permutation group $\mathrm{S}_n$,  and the Rubik's cube group can be produced by a finite set of generators. Even for large groups, the number of generators is \emph{much} smaller than the size of the group: $1$ for $\mathbb{Z}_n$ of size $n$, $2$ for $\mathrm{S}_n$ of size $n!$, and $6$ for the cube group of size $4\times 10^{19}$.

\textbf{Continuous Groups and Infinitesimal Generators.}
Similarly, Lie theory provides a way of analyzing continuous groups in terms of their \emph{infinitesimal} generators. The Lie Algebra $\mathfrak{g}$ of a Lie Group $G$ (a continuous group that forms a smooth manifold) is the tangent space at the identity $\mathfrak{g}:=T_{\mathrm{id}}G \subseteq \mathbb{R}^{n\times n}$, which is a vector space of infinitesimal generators of group transformations from $G$. The exponential map $\mathrm{exp}: \mathfrak{g} \to G$ maps back to the Lie Group and can be understood through the series: $\mathrm{exp}(A) = \sum_{k=0}^\infty A^k/k!$

A classic example is the rotation group $G=\mathrm{SO}(n)$ with matrices $\mathbb{R}^{n\times n}$ satisfying $R^\top R=I$ and $\mathrm{det}(R)=1$. Parametrizing a curve $R(t)$ with $R(0)=I$, $R'(0)=A$, one can find the tangent space by differentiating the constraint at the identity. The Lie Algebra consists of antisymmetric matrices: $\mathfrak{so}(n) = T_\mathrm{id}\mathrm{SO}(n) = \{A\in \mathbb{R}^{n\times n}: A^\top=-A \}$.

Given that $\mathfrak{g}$ is a finite dimensional vector space ($D=\mathrm{dim}(\mathfrak{g})=\mathrm{dim}(G)$), its elements can be expanded in a basis $\{A_1,A_2,...,A_D\}$. For some Lie Groups like $\mathrm{SO}(n)$, the orientation preserving isometries $\mathrm{SE}(n)$, the special unitary group $\mathrm{SU}(n)$, the symplectic group $\mathrm{Sp}(n)$, the exponential map is surjective meaning all elements $g\in G$ can be written in terms of this exponential $g=\mathrm{exp}(\sum_i\alpha_iA_i)$ with a set of real valued coefficients $\{\alpha_i\}_{i=1}^D$.
But in general for other Matrix Groups like $\mathrm{O}(n)$, $\mathrm{E}(n)$, and $\mathrm{O}(1,3)$, $\mathrm{exp}$ is not surjective and one can instead write $g = \mathrm{exp}(\sum_i\alpha_iA_i)\Pi_{i=1}^Nh_{k_i}$ as a product of the exponential map that traverses the identity component and an additional collection of discrete generators (see \autoref{sec:app_finitely_generated}).

\textbf{Group Representations.}
In the machine learning context, a group element is most relevant in how it acts as a transformation on an input. A (linear finite dimensional) group \emph{representation} $\rho: G \to \mathrm{GL}(m)$ associates each $g\in G$ to an invertible matrix $\rho(g)\in \mathbb{R}^{m\times m}$ that acts on $\mathbb{R}^m$. The representation satisfies $\forall g_1,g_2\in G: \rho(g_1g_2)=\rho(g_1)\rho(g_2)$, and therefore also $\rho(g^{-1})=\rho(g)^{-1}$. The representation specifies how objects transform under the group, and can be considered a specification of the \emph{type} of an object.

\textbf{Lie Algebra Representations.}
Mirroring the group representations, Lie Groups have an associated representation of their Lie algebra, prescribing how infinitesimal transformations act on an input. A Lie algebra representation $d\rho: \mathfrak{g} \to \mathfrak{gl}(N)$ is a \textbf{linear} map from the Lie algebra to ${m\times m}$ matrices. 
An important result in Lie Theory relates the representation of a Lie Group to the representation of its Lie Algebra 
\begin{equation}\label{eq:lie_correspondance}
\forall A \in \mathfrak{g}: \quad \rho(e^A) = e^{d\rho(A)}
\end{equation}

\textbf{Tensor Representations.} 
Given some base group representation $\rho$, Lie Algebra representation $d\rho$, acting on a vector space $V$, 
representations of increasing size and complexity can be built up through the tensor operations dual ($*$), direct sum ($\oplus$), and tensor product ($\otimes$).

\begin{center}
    \begin{tabular}{cccc}
         \toprule
         OP & $\rho$ &  $d\rho$& $V$ \\
         \midrule
         * & $\rho(g^{-1})^{\top}$ & $-d\rho(A)^\top $& $V^*$\\
         $\oplus$ & $\rho_1(g)\oplus \rho_2(g) $&$d\rho_1(A)\oplus d\rho_2(A)$& $V_1\oplus V_2$\\
         $\otimes$ &$\rho_1(g)\otimes \rho_2(g)$&$d\rho_1(A)\bar{\oplus} d\rho_2(A)$ &$V_1\otimes V_2$\\
         \bottomrule
    \end{tabular}
\end{center}

Acting on matrices, $\oplus$ is the direct sum which concatenates the matrices on the diagonal $X\oplus Y = \begin{bmatrix} X & 0 \\ 0  & Y \\\end{bmatrix}$, and facilitates multiple representations which are acted upon separately. The $\otimes$ on matrices is the Kronecker product, and $\bar{\oplus}$ is the \emph{Kronecker sum}: $X\bar{\oplus} Y = X \otimes I +I \otimes Y$. $V^*$ is the dual space of $V$. The tensor product and dual are useful in describing linear maps from one vector space to another. Linear maps from $V_1\to V_2$ form the vector space $V_2\otimes V_1^*$ and have the corresponding representation $\rho_2 \otimes \rho_1^*$. 

We will work with the corresponding vector spaces and representations interchangeably with the understanding that the other is defined through these composition rules. We abbreviate many copies of the same vector space $\underbrace{V\oplus V\oplus ... \oplus V}_{m}$ as $mV$. Similarly we will refer to the vector space formed from many tensor products $T_{(p,q)} = V^{\otimes p} \otimes (V^*)^{\otimes q }$ where $(\cdot)^{\otimes p}$ is the tensor product iterated $p$ times. Following the table, these tensors have the group representation $\rho_{(p,q)}(g) = \rho(g)^{\otimes p} \otimes \rho^*(g)^{\otimes q}$, and the Lie algebra representation $d\rho_{(p,q)}(A) = d\rho(A)^{\bar{\oplus} p} \bar{\oplus} d\rho^*(A)^{\bar{\oplus} q}$. We will abbreviate $T_{p+q}$ for $T_{(p,q)}$ when using orthogonal representations ($\rho=\rho^*$), as the distinction between $V$ and $V^*$ becomes unnecessary.

\section{Equivariant Linear Maps}
In building equivariant models, we need that the layers of the network are equivariant to the action of the group. Below we characterize all equivariant linear layers $W\in \mathbb{R}^{N_2\times N_1}$ that map from one vector space $V_1$ with representation $\rho_1$ to another vector space $V_2$ with representation $\rho_2$ for a matrix group $G$. We prove that the infinite set of constraints can be reduced to a finite collection without loss of generality, and then provide a polynomial-time algorithm for solving the constraints.
\subsection{The Equivariance Constraint}
Equivariance requires that transforming the input is the same as transforming the output:
\begin{equation*}
    \forall x \in V_1, \forall g\in G: \qquad \rho_2(g)Wx=W\rho_1(g)x.
\end{equation*}
Since true for all $x$, $\rho_2(g)W\rho_1(g)^{-1}=W$, or more abstractly:
\begin{equation}\label{eq:expensive_equivariance}
    \forall g\in G:\quad \rho_2(g)\otimes \rho_1(g^{-1})^\top \mathrm{vec}(W)=\mathrm{vec}(W)
\end{equation}
where $\mathrm{vec}$ flattens the matrix into a vector. $\rho_1(g^{-1})^\top$ is the dual representation $\rho_1^*(g)$, and so the whole object $\rho_2(g)\otimes \rho_1(g^{-1})^\top = (\rho_2\otimes \rho_1^*)(g)=\rho_{21}(g)$ is a representation of how $g$ acts on matrices mapping from $V_1\to V_2$.

While equation \eqref{eq:expensive_equivariance} is linear, the constraint must be upheld for each of the possibly combinatorially large or infinite number of group elements in the case of continuous groups. However, in the following section we show that these constraints can be reduced to a finite and small number.

\subsection{General Solution for Symmetric Objects}
Equation \eqref{eq:expensive_equivariance} above with $\rho=(\rho_2\otimes \rho_1^*)$ is a special case of a more general equation expressing the symmetry of an object $v$,
\begin{equation}\label{eq:symmetry}
    \forall g \in G: \quad \rho(g)v=v
\end{equation}

Writing the elements of $G$ in terms of their generators: $g = \mathrm{exp}(\sum_i^D\alpha_iA_i)\Pi_{i=1}^Nh_{k_i}$.
For group elements with $k = \varnothing$, we have
\begin{equation*}
    \forall \alpha_i:\quad \rho\big(\exp({\sum_i\alpha_iA_i})\big)v = v
\end{equation*}
Using the Lie Algebra - Lie Group representation correspondence \eqref{eq:lie_correspondance} and the linearity of $d\rho(\cdot)$ we have
\begin{equation*}
    \forall \alpha_i:\quad \exp\big({\sum_i\alpha_id\rho(A_i)}\big)v = v.
\end{equation*}
Taking the derivative with respect to $\alpha_i$ at $\alpha=0$, we get a constraint for each of the infinitesimal generators
\begin{equation}\label{eq:constraint_lie}
    \mywboxmath{\forall i=1,...,D: \quad d\rho(A_i)v = 0}
\end{equation}
For group elements with all $\alpha_i=0$ and $N=1$, we get an additional constraint for each of the discrete generators in the group:
\begin{equation}\label{eq:constraint_discrete}
    \mywboxmath{\forall k=1,...,M: \quad (\rho(h_k)-I)v = 0}.
\end{equation}

We get a total of $O(M+D)$ constraints, one for each of the discrete and infinitesimal generators.
In \autoref{sec:app_constraints_proof}, we prove that these reduced constraints are not just \textbf{necessary} but also \textbf{sufficient}, and therefore characterize all solutions to the symmetry equation \eqref{eq:symmetry}.

\textbf{Solving the Constraint:}
We collect each of the symmetry constraints $C_1=d\rho(A_1), C_2=d\rho(A_2),...,C_{D+1}=\rho(h_1)-I,...$ into a single matrix $C$, which we can breakdown into its nullspace spanned by the columns of $Q\in \mathbb{R}^{m\times r}$ and orthogonal complement $P \in \mathbb{R}^{m\times (m-r)}$ using the singular value decomposition: 
\begin{align}\label{eq:svd}
    Cv = \begin{bmatrix} d\rho(A_1) \\
d\rho(A_2)\\ ... \\\rho(h_1)-I\\ ...\end{bmatrix}v = U\begin{bmatrix} \Sigma & 0 \\ 0 & 0\\\end{bmatrix}\begin{bmatrix}P^\top \\ Q^\top \end{bmatrix} v=0.
\end{align} 

All symmetric solutions for $v$ must lie in the nullspace of $C$: $v = Q\beta$ for some coefficients $\beta$, and we can then parametrize all symmetric solutions directly in this subspace. Alternatively, defining $\beta = Q^Tv_0$ we can reuse any standard parametrization and initialization, but simply project onto the equivariant subspace: $v = QQ^Tv_0$.

Thus given any finite dimensional linear representation, we can solve the constraints with a singular value decomposition.\footnote{Equations \eqref{eq:constraint_lie} and \eqref{eq:constraint_discrete} apply also to infinite dimensional representations, where $\rho$ and $d\rho$ are linear operators acting on functions $v$, but solving these on a computer would be more difficult.}. If $v\in \mathbb{R}^m$ the runtime of the approach is $O((M+D)m^3)$.

\begin{figure}[!t]
    \centering
    \subfigure[$S_4$]{\includegraphics[width=.23\linewidth]{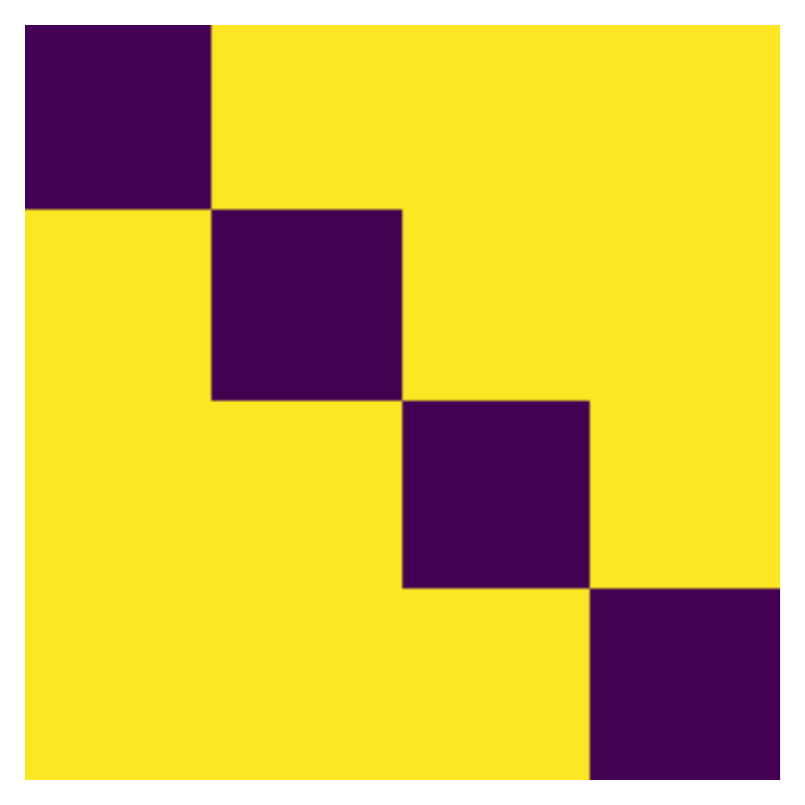}}
    \subfigure[$\mathbb{Z}_4$]{\includegraphics[width=.23\linewidth]{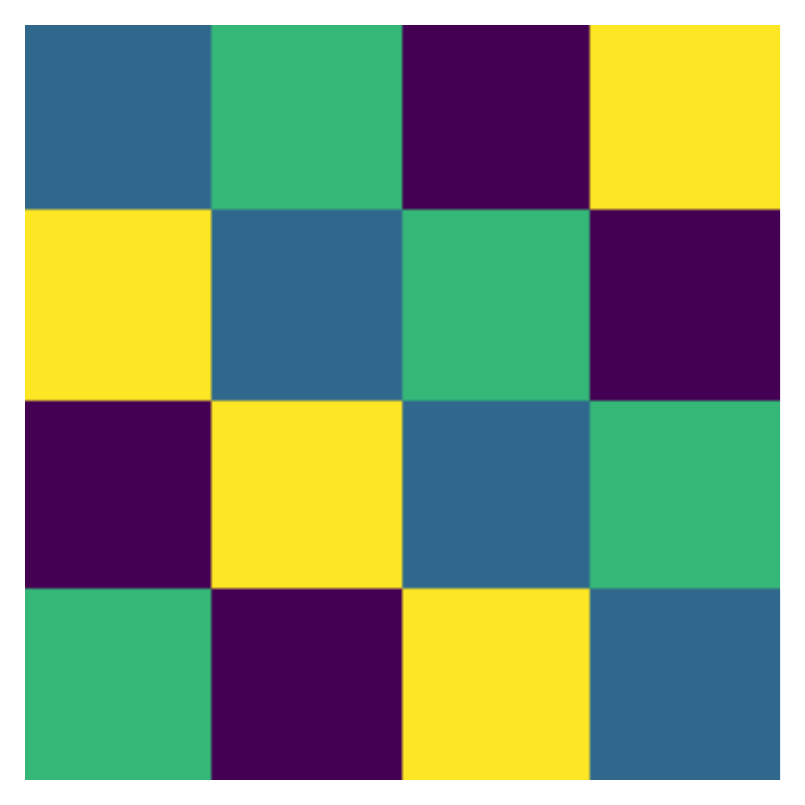}}
    \subfigure[$\mathbb{Z}_2^2$]{\includegraphics[width=.23\linewidth]{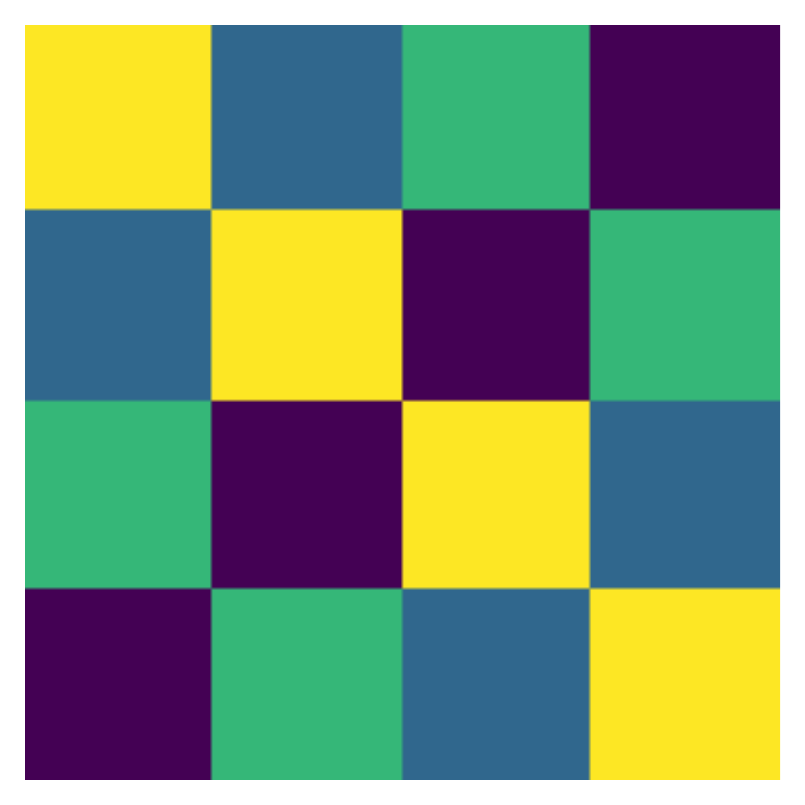}}
    \subfigure[$\mathbb{Z}_4 \ltimes \mathbb{Z}_2^2$]{\includegraphics[width=.23\linewidth]{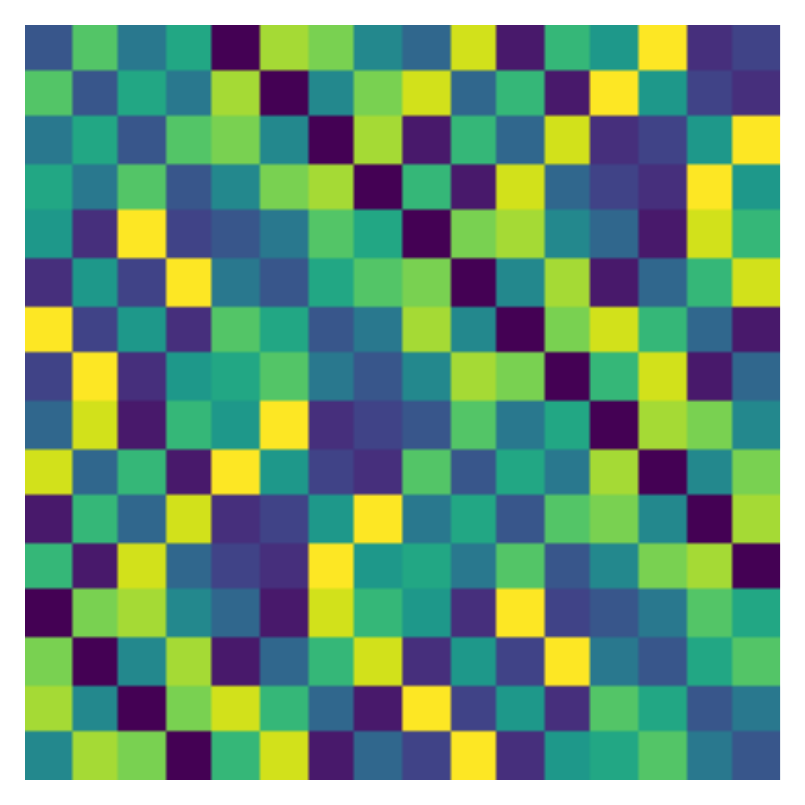}}
    \caption{Equivariant basis for permutations, translation, 2d translation, and GCNN symmetries respectively, each of which are solutions to \autoref{eq:constraint_discrete} for different groups. The $r$ different solutions in the basis are shown by different colors.}
    \label{fig:solns_visualized}
\end{figure}

\subsection{A Unifying Perspective on Equivariance}
In order to make it more concrete and demonstrate its generality, we now show that standard convolutional layers \citep{lecun1989backpropagation}, deep sets \citep{zaheer2017deep}, invariant graph networks \citep{maron2018invariant}, and GCNNs \citep{cohen2016group} are examples of the solutions in equation \eqref{eq:svd} when specifying a specific symmetry group and representation.

\textbf{Convolutions}:
To start off with the 1D case with sequences of $n$ elements and a single channel, $V=\mathbb{R}^n$ is acted upon by cyclic translations from the group $G=\mathbb{Z}_n$. The group can be generated by a single element given by the permuation matrix $\rho(h) = \mathrm{P}[n,1,2,...,n-1]$. Equivariant linear maps from $V\to V$ are of type $T_{(1,1)}$. Expressing the representation and solving eq. \eqref{eq:svd} with SVD gives the $r=n$ matrices (reshaped from the rows of $Q$) shown by the circulant matrix in \autoref{fig:solns_visualized}, which is \emph{precisely} the way to express convolution as a matrix.

In the typical case of 2D arrays with $V = \mathbb{R}^{n^2}$ elements and multiple channels $c_{\mathrm{in}}$ $c_{\mathrm{out}}$, there are $M=2$ generators of the group $G=\mathbb{Z}_n \times \mathbb{Z}_n = \mathbb{Z}_n^2$ that are $\rho(h_1) = \rho(h)\otimes I$ and $\rho(h_2) = I\otimes \rho(h)$ defined in terms of the generator in the 1D case. For multiple channels, the mapping is $c_{\mathrm{in}}V\to c_{\mathrm{out}}V$ which has type $c_{\mathrm{in}}c_{\mathrm{out}}T_{(1,1)}$ which yields the matrix valued 2D convolution (with $c_{\mathrm{in}}c_{\mathrm{out}}n^2$ independent basis elements) that we are accustomed to using for computer vision\footnote{Note that the inductive bias of \emph{locality} restricting from $n\times n$ filters to $3\times 3$ filters is not a consequence of equivariance.}.

\textbf{Deep Sets}:
We can recover the solutions in \citet{zaheer2017deep} by specifying $V=\mathbb{R}^n$ and considering $S_n$ (permutation) equivariant linear maps $V\to V$. $S_n$ can be generated in several ways such as with the $M=2$ generators $\rho(h_1) = \mathrm{P}[1,n-1,2,3,...]$ and $\rho(h_2) = \mathrm{P}[2,1,3,4,...]$ \citep{conrad2013generating}. Solving the constraints for $T_{(1,1)}$ yields the $r=2$ dimensional basis $Q = [I,\mathds{1}\mathds{1}^\top]$ shown in  \autoref{fig:solns_visualized}.

\textbf{Equivariant Graph Networks}:
Equivariant graph networks in \citet{maron2018invariant} generalize deep sets to $S_n$ equivariant maps from $T_k \to T_\ell$, such as maps from adjacency matrices $T_2$ to themselves. They show these maps satisfy 
\begin{equation}\label{eq:maron}
    \forall P \in S_n: \ P^{\otimes (k+\ell)}\mathrm{vec}(W)=\mathrm{vec}(W),
\end{equation}
and use analytic techniques to find a basis, showing that the size of the basis is upper bounded\footnote{For small $n$, the size of the equivariant basis for $T_k$ can actually be less than $B_k$ when $n^k<B_k$.} by the Bell numbers $1,2,5,15,...$. Noting that $P=(P^{-1})^\top$, we can now recognize $P^{\otimes k+\ell} = \rho_{(k,\ell)}(P)$ acting on the maps of type $T_{(k,\ell)}$. However we need not solve the combinatorially large \autoref{eq:maron}; our algorithm instead solves it just for the permutation generators $\rho(h_i)$, yielding the same solutions.

\textbf{GCNNs}: The Group Equivariant CNNs in \citet{cohen2016group} can be defined abstractly through fiber bundles and base spaces, but we can also describe them in our tensor notation. The original GCNNs have the $G = \mathbb{Z}_4 \ltimes (\mathbb{Z}_n\times \mathbb{Z}_n)$ symmetry group consisting of discrete translations of the grid, as well as $90^o$ rotations where $\ltimes$ is the semi-direct product. \footnote{\citet{cohen2016group} also make $\mathrm{D}_4$ dihedral equivariant networks that respect reflections, which can be accomodated by in our framework with $1$ additional generator.} In total, the representation space can be written $V = \mathbb{R}^4\otimes \mathbb{R}^{n^2}$. We can now disentangle these two parts to read off the $M=3$ generators for $x$, $y$ translation and rotation. The translation generators are $I\otimes \rho(h_1)$ and $I\otimes \rho(h_2)$ from the 2D convolution section, as well a generator for rotation $P[4,1,2,3] \otimes \mathrm{Rot}_{90}$ with the $\mathrm{Rot}_{90}$ matrix performing $90^o$ rotations on the grid. Solving for the constraint on $T_{(1,1)}$ yields the $G$-convolutional layer embedded in a dense matrix shown in \autoref{fig:solns_visualized}. Note the diagonal blocks implement rotated copies of a given filter, equivalent to the orientations in the regular representation of a GCNN.

\begin{figure}[!t]
    \centering
    \subfigure[$T_4^{S_{6}}$]{\includegraphics[width=.32\linewidth]{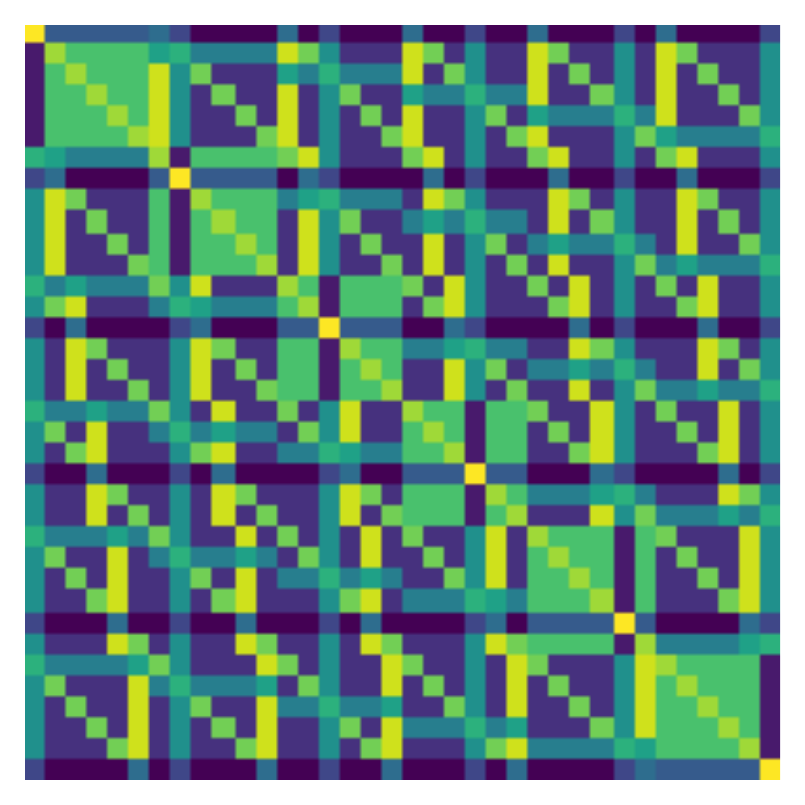}}
    \subfigure[ $T_2^\mathrm{Rubiks}$]{\includegraphics[width=.32\linewidth]{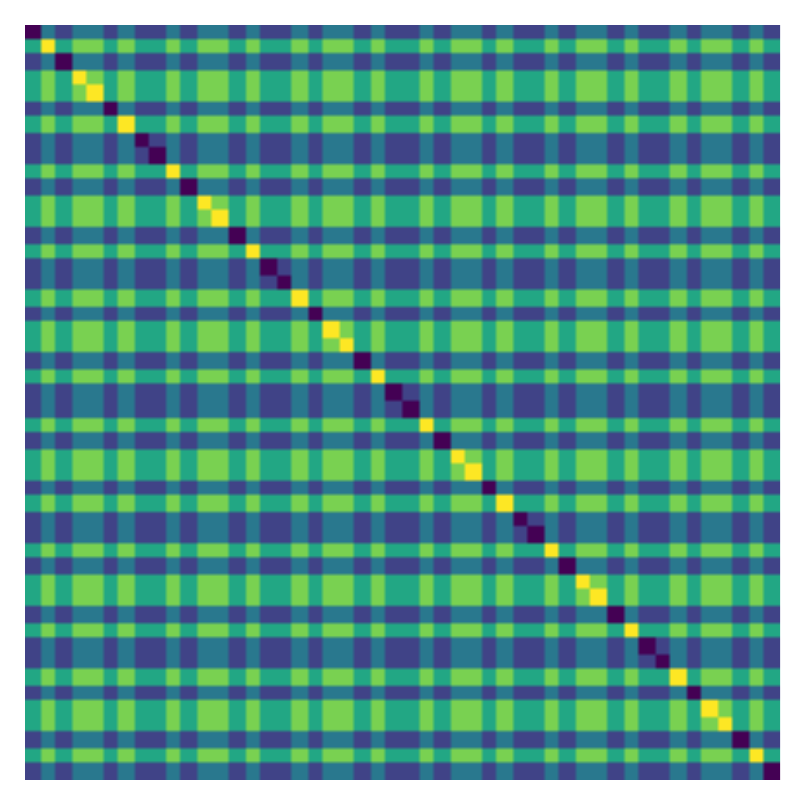}}
    \subfigure[$T_4^{\mathrm{SO}(3)}$]{\includegraphics[width=.32\linewidth]{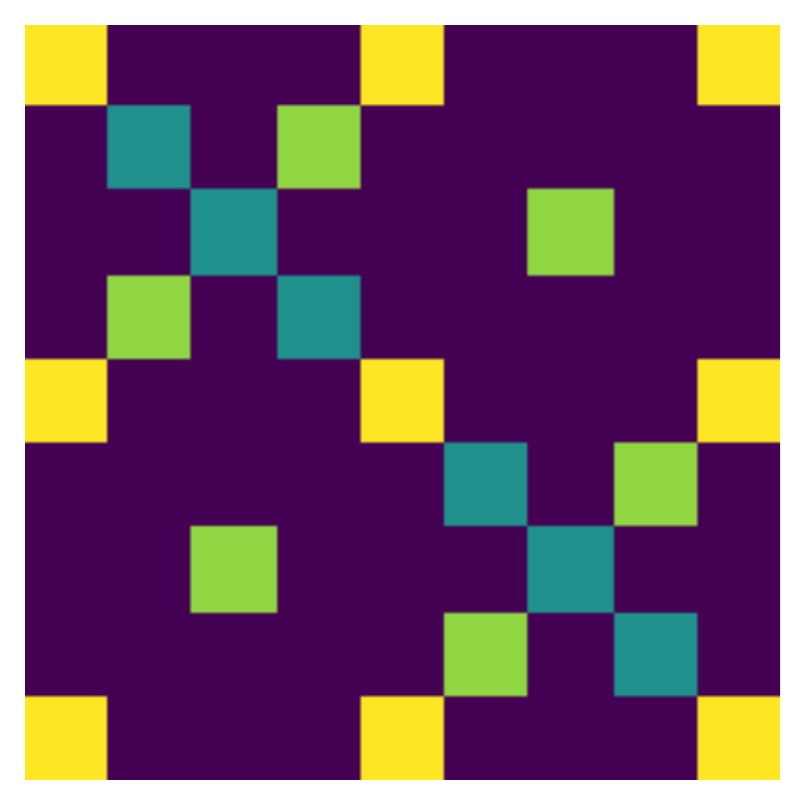}}
    \caption{Equivariant basis for various tensor representations $T_k^G$ where $G$ denotes the symmetry group. The $r$ different solutions in the basis are shown by different colors. For $\mathrm{SO}(3)$ the bases cannot be separated into disjoint set of $0$ or $1$ valued vectors, and so we choose overlapping colors randomly and add an additional color for $0$.}
    \label{fig:app_bases}
\end{figure}

Notably, each of these solutions for convolution, deep sets, equivariant graph networks, and GCNNs are produced as solutions from \autoref{eq:svd} as a direct consequence of specifying the representation and the group generators. In \autoref{sec:app_bases} we calculate the equivariant basis for tensor representations of these groups $\mathbb{Z}_n,\mathrm{S}_n$, $\mathrm{D}_n$, as well as unexplored territory with $\mathrm{SO}(n)$, $\mathrm{O}(n)$, $\mathrm{Sp}(n)$, $\mathrm{SO}^+(1,3)$, $\mathrm{SO}(1,3)$, $\mathrm{O}(1,3)$, $\mathrm{SU}(n)$, and the Rubiks Cube group. We visualize several of these equivariant bases in \autoref{fig:app_bases}.

\FloatBarrier

\section{Efficiently Solving the Constraint}
The practical application of our general approach is limited by two factors: the computational cost of computing the equivariant basis at initialization, and the computational cost of applying the equivariant maps in the forward pass of a network. In this section we address the scalability of the first factor, computing the equivariant basis.

The runtime for using SVD directly to compute the equivariant basis is too costly for all but very small representations $m = \mathrm{dim}(V) < 5000$. We improve upon the naive algorithm with two techniques: dividing the problem into a smaller set of independent subproblems and exploiting structure in the constraint matrices to enable an efficient iterative Krylov subspace approach for computing the nullspace. These two techniques allow us to compute the bases for high dimensional representations while not sacrificing the equivariance or completeness of the solution basis. Our resulting networks run in time similar to a standard MLP.

\subsection{Dividing into Independent Sub-problems}

The feature space $U$ in a neural network can be considered a combination of objects with different types and multiplicities. The features in standard CNN or deep set would be $c$ copies of rank one tensors, $U = cT_1$, where $c$ is the number of channels. Graph networks include both node features $T_1$ as well as edge features $T_2$ like the adjacency matrix. More general networks could have a mix of representations, for example $100$ scalars, $30$ vectors, $10$ matrices and $3$ higher order tensors: $U = 100T_0\oplus 30T_1\oplus 10T_2\oplus 3T_3$. These composite representations with multiplicity are built from direct sums of simpler representations. $\rho_U(g) =\bigoplus_{a\in \mathcal{A}} \rho_a(g)$ for some collection of representations $\mathcal{A}$. 

Since linear maps $U_1 \to U_2$ have the representation $\rho_{2} \otimes \rho_{1}^*$, the product can be expanded as the direct sum

\begin{equation}\label{eq:subproblems}
    \rho_{2} \otimes \rho_{1}^* = \bigoplus_{b\in \mathcal{A}_2} \rho_{b} \otimes \bigoplus_{a\in \mathcal{A}_1} \rho^*_{a} = \bigoplus_{(b,a) \in   \mathcal{A}_2\times \mathcal{A}_1} \rho_b \otimes \rho_a^*.
\end{equation}

Since $\oplus$ for both the group and algebra representations concatenates blocks along the diagonal, the constraints can be separated into the blocks given by each of the $(b,a)$ pairs. Each of these constraints \eqref{eq:constraint_lie} and \eqref{eq:constraint_discrete} can be solved independently for the $\rho_b \otimes \rho_a^*$ representation and then reassembled into the parts of the full matrix.

Unlike Steerable CNNs which use analytic solutions of irreducible representations \citep{cohen2016steerable,weiler2019general}, we need not worry about any Clebsch-Gordon coefficients or otherwise, regardless of the representation used. \footnote{For irreducible representations one typically decomposes $\rho_i \otimes \rho_j = Q^{-1}\big(\bigoplus_k \rho_k\big) Q$ with Clebsch-Gordan matrix $Q$, but we can leave the rep as $\rho_i \otimes \rho_j$ and solve numerically.} Note that tensor representations make things especially simple since $T_{(p,q)}\otimes T_{(r,s)}^* = T_{(p+s,q+r)}$, but are not required.

\subsection{Krylov Method for Efficient Nullspaces}
We can exploit structure in the matrices $\rho$ and $d\rho$ for a more efficient solution.
  With this in mind, we propose to find the nullspace $Q\in \mathbb{R}^{n\times r}$ where $r$ is the rank of the nullspace with the following optimization problem:
\begin{equation}
    \mathrm{min} \ \  \|CQ\|_F^2 \quad s.t \quad Q^\top Q=I.
\end{equation}
Minimizing using gradient descent, we have a very close relative of QR power iteration \citep{francis1961qr} and Oja's rule \citep{garber2015fast,de2015global,shamir2015stochastic}, that instead finds the smallest singular vectors. As the nullspace components are preserved by the gradient updates, the orthogonalization constraint can in fact be removed during the minimization and we list the steps of the iterative method in \autoref{alg:subspace}. Crucially, gradients require only matrix vector multiplies (MVMs) with the constraint matrix $C$, we never have to form the representation matrices explicitly and can instead implement an efficient MVM for $\rho$ and $d\rho$. Through iterative doubling of the max rank $r$ we need not know the true rank beforehand. As we prove in \autoref{sec:krylov_runtime} the algorithm produces an $\epsilon$ accurate solution in time $O((M+D)\mathcal{T}r\log(1/\epsilon)+r^2n)$ where $\mathcal{T}$ is the time for an MVM with $\rho$ and $d\rho$. The method is "exact" in the sense of numerical algorithms in that we can specify a precision $\epsilon$ close to machine precision and converge in $\log(1/\epsilon)$ iterations due to the exponential convergence rate, which we verify in \autoref{fig:convergence}.
\begin{figure}[t]
    \centering
	\includegraphics[width=0.47\textwidth]{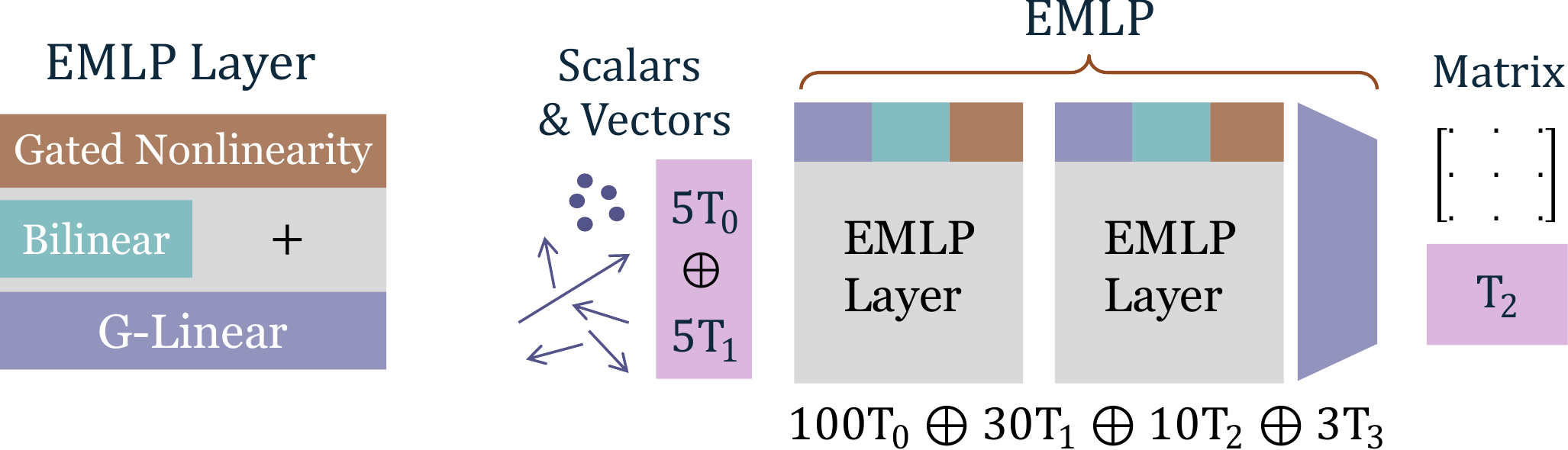}
	\caption{EMLP layers. G-equivariant linear layers, followed by the bilinear layer and a shortcut connection, and finally a gated nonlinearity. Stacking these layers together and choosing some internal representation (shown below), the EMLP maps some collection of geometric quantities to some other collection. Here we show the equivariant mappings from scalars and vectors to matrices.
	}
    \label{fig:architecture}
    \vspace{-.3cm}
\end{figure}
\begin{algorithm}
\SetAlgoLined
def \textbf{KrylovNullspace}($C$):\\
$r_\mathrm{max} = r = 10$\\
\While{$r = r_\mathrm{max}$}{
$r_\mathrm{max} \gets 2r_\mathrm{max}$\\
$Q = \mathrm{CappedKrylovNullspace}(C,r_\mathrm{max})$\\
$r \gets \mathrm{rank}(Q)$
}
\Return $Q$\\
def \textbf{CappedKrylovNullspace}($C,r_\mathrm{max}$):\\
 $Q \sim \mathcal{N}(0,1)^{n\times r_\mathrm{max}}$\\
\While{$L(Q) > \epsilon$}{
    $L(Q) = \|CQ\|^2_F$\\
    $Q \gets Q - \eta\nabla L$
}
$Q,\Sigma,V = \mathrm{SVD}(Q)$\\
\Return $Q$\\
\caption{Fast Krylov Nullspace}
\label{alg:subspace}
\end{algorithm}

The pairs of tensor products of representations, $\rho_b(h)\otimes \rho_a(h^{-1})^\top$ and $d\rho_b(A)\bar{\oplus} (-d\rho_a(A)^\top)$ from \autoref{eq:subproblems} have Kronecker structure allowing efficent MVMs $(A\otimes B) \mathrm{vec}(W) = \mathrm{vec}(AWB^\top)$. Exploiting this structure alone, solving the constraints for a matrix $W \in \mathbb{R}^c\to\mathbb{R}^c$ takes time 
\begin{equation}
    O((M+D)\mathcal{T}rc+r^2c^2)
\end{equation} where $\mathcal{T}$ is the time for MVMs with constituent matrices $\rho_a,\rho_b,d\rho_a,d\rho_b$. For some of the groups this time $\mathcal{T}$ is in fact a \emph{constant}, for example the permutation generators merely swap two entries, and Lie algebras can often be written in a sparse basis. For high order tensor representations, one can exploit higher order Kronecker structure. 
Even for discrete groups the runtime is a strict improvement over the approach by \citet{van2020mdp} which runs in time $O(|G|\mathcal{T}rc+r^2c^2)$. For large discrete groups like $S_n$ our approach gives an exponential speedup, $O(n!)\rightarrow O(n)$.

\begin{figure*}[!ht]
	\centering
	\subfigure[O(5) invariant Synthetic]{
		\includegraphics[width=0.335\textwidth]{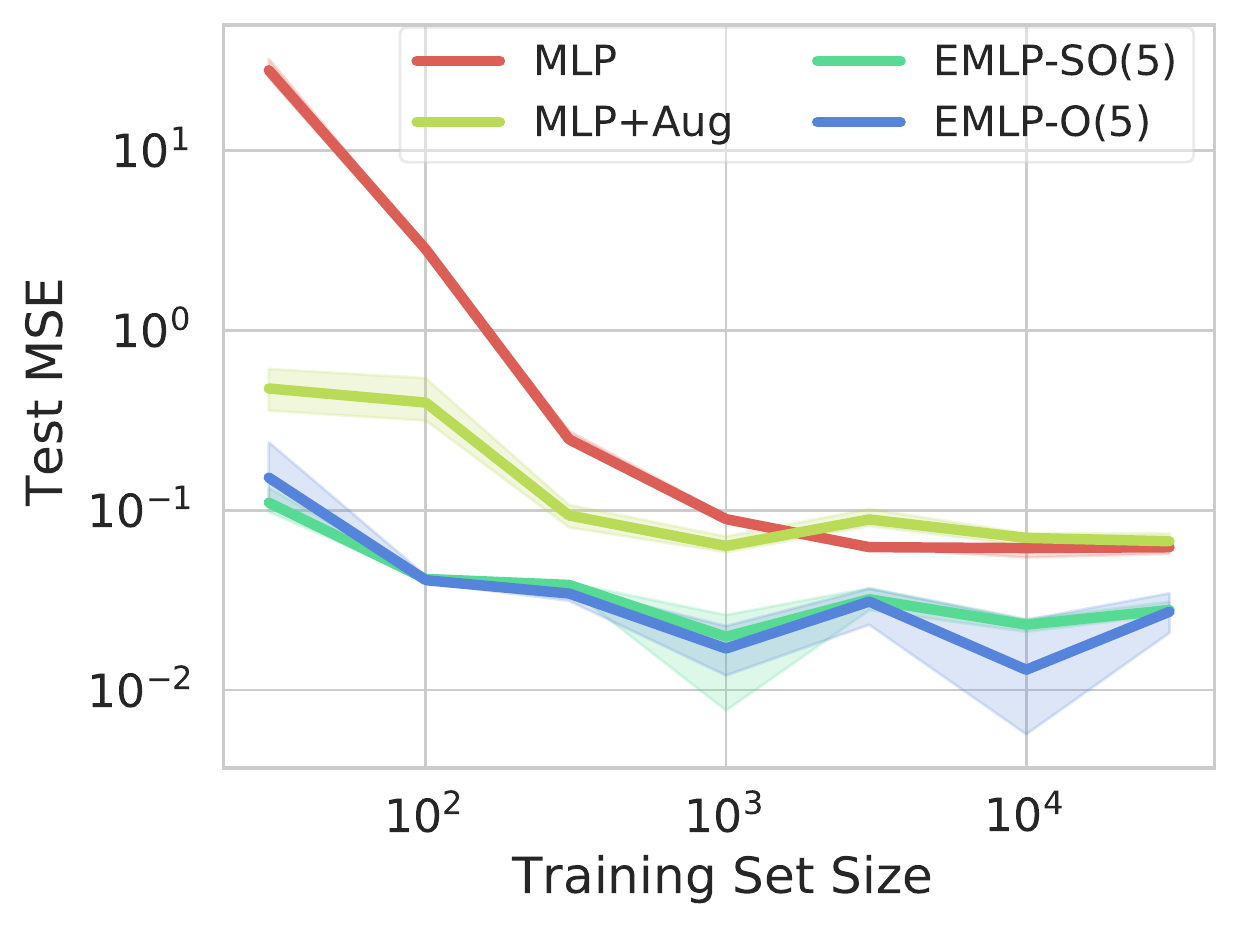}
	}
	\subfigure[O(3) equivariant Inertia]{
		\includegraphics[width=0.32\textwidth]{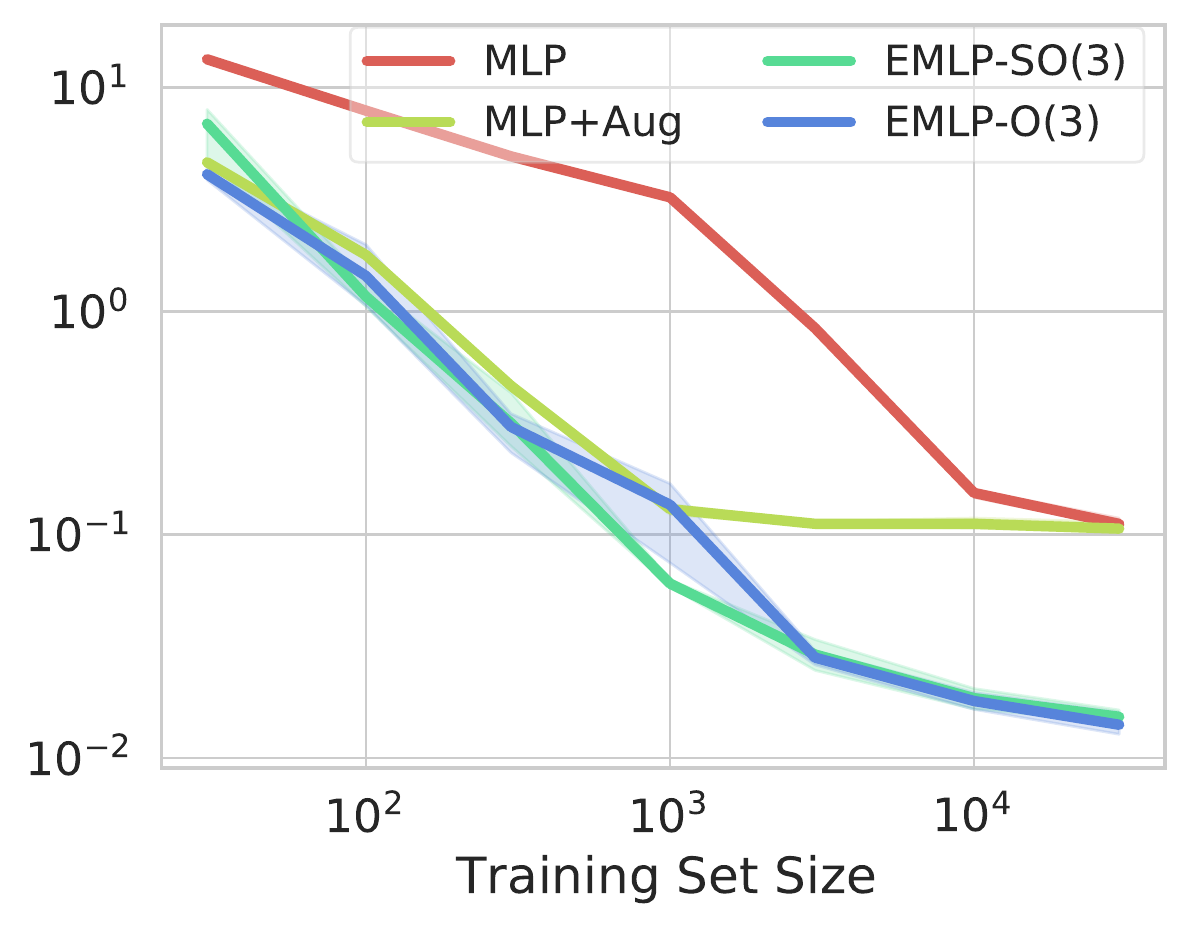}
	}
	\subfigure[O(1,3) invariant $(e^- + \mu^-)$-Scattering]{
		\includegraphics[width=0.32\textwidth]{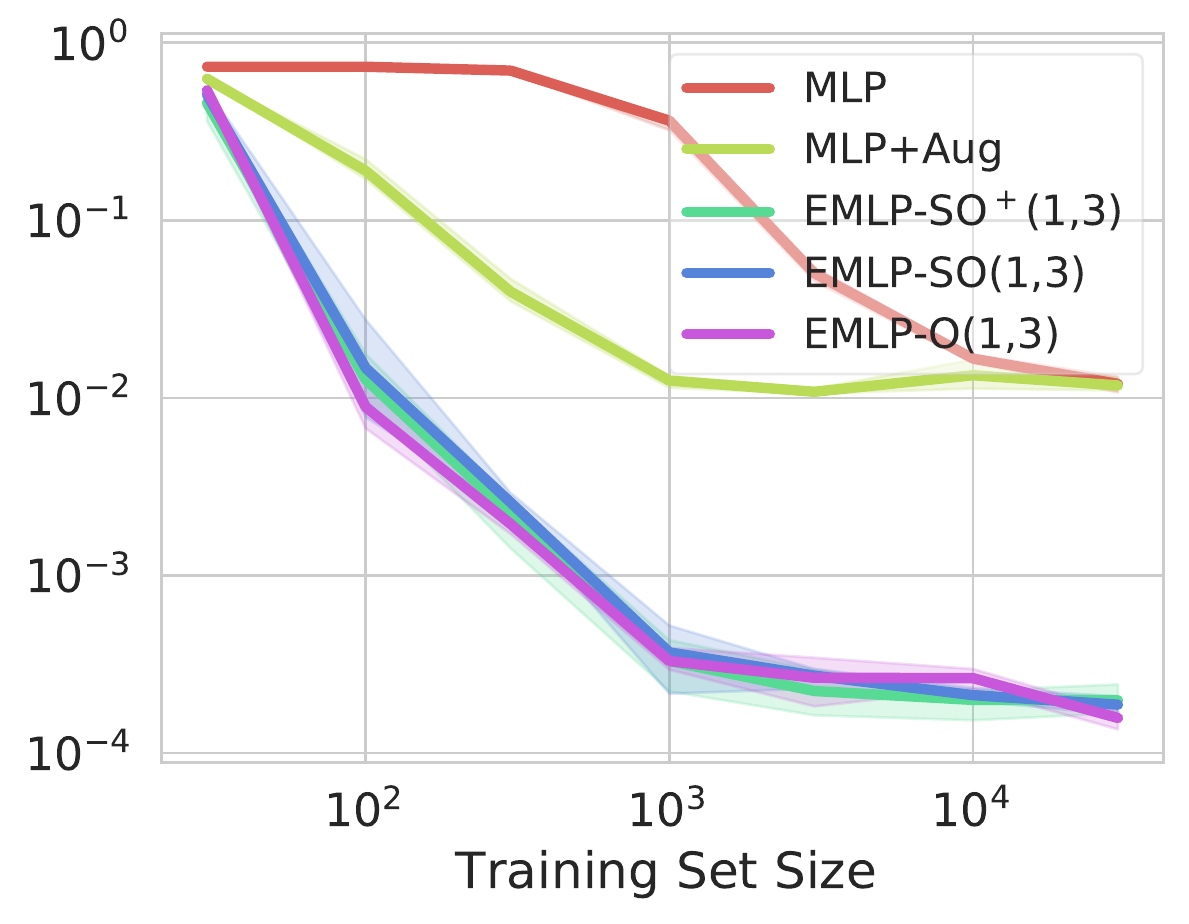}
	}
	\caption{
		Data efficiency for the synthetic equivariance experiments. Here the EMLP-$G$ models where $G$ are relevant symmetry groups strongly outperforms both standard MLPs and MLPs that have been trained with data augmentation to the given symmetry group, across the range of dataset sizes. The shaded regions depict $95\%$ confidence intervals taken over $3$ runs.
        }
        \label{fig:data_efficiency}
\end{figure*}

\section{Network Architecture}
While the constraint solving procedure can be applied to any linear representations, we will use tensor representations to construct our network. The features in each layer are a collection of tensors of different ranks $v \in U = \bigoplus_{a\in \mathcal{A}} T_{(p_a,q_a)}$ with the individual objects $v_a \in T_{(p_a,q_a)}$. As a heuristic, we allocate the channels uniformly between tensor ranks for the intermediate layers. For example with $256$ channels for an $\mathrm{SO}(3)$ equivariant layer with $\mathrm{dim}(T_{(p,q)})=3^{p+q}$, uniformly allocating channels produces $U=70T_0\oplus23T_1+7T_2+2T_3$. The input and output layers are set by the types of the data. To build a full equivariant multilayer perceptron (EMLP) from the equivariant linear layer, we also need equivariant nonlinearities. 

\textbf{Gated Nonlinearities}:
For this purpose we use \emph{gated} nonlinearities introduced in \citet{weiler20183d}. Gated nonlinearities act separately for each of the different objects in the features (that are concatenated through the direct sum $z = \mathrm{Concat} (\{v_a\}_{a \in \mathcal{A}})$). 
The nonlinearity takes values $\mathrm{Gated}(v_a) = v_a\sigma(s_a)$ where $s_a$ is a scalar 'gate' for each of the objects. For scalar objects $v_a\in T_0=\mathbb{R}^1$ and regular representations (which allow pointwise nonlinearities), the gate is just the object itself and so the nonlinearity is just Swish \citep{ramachandran2017searching}. For other representations the gate scalars are produced as an additional output of the previous layer.

\textbf{Universality}:
The theorem in \citet{maron2019universality} shows that tensor networks with pointwise nonlinearities and $G$-equivariant linear layers for $G\le S_n$ are universal. However, this result does not extend to the gated nonlinearities required for other groups and representations. As we prove in \autoref{sec:app_bilinear_necessary}, gated nonlinearities are \emph{not} sufficient for universality in this general case, and can be extremely limiting in practice. The problem relates to not being able to express any kind of contractions between elements with the different objects within a feature layer (like a dot product).

\textbf{Cheap Bilinear Layers.} To address this limitation we introduce an inexpensive bilinear layer which performs tensor contractions on pairs of input objects that produce a given output type. Explicitly, two input objects $v_a \in T_{(a_1,a_2)}$ and $v_b \in T_{(b_1,b_2)}$ can be contracted to give a type $T_{(c_1,c_2)}$ if and only if $(a_1,a_2) = (c_1+b_2,c_2+b_1)$ or $(b_1,b_2) = (c_1+a_2,c_2+a_1)$. In other words, if $v_a$ can be interpreted as a linear map from $T_b\rightarrow T_c$ then we can apply $y_c = \mathrm{Reshape}(v_a)v_b$ and vice versa. We add a learnable parameter weighting each of these contractions (excluding scalars).

We can now assemble the components to build a full equivariant multilayer perceptron (EMLP) from the equivariant linear layer, the gated nonlinearities, and the additional bilinear layer. We show how these components are assembled in \autoref{fig:architecture}.

\section{Experiments}
We evaluate EMLP on several synthetic datasets to test its capability on previously unexplored groups, and apply our model to the task of learning dynamical systems with symmetry.

\subsection{Synthetic Experiments}
\textbf{O(5) Invariant Task}:
To start off, we evaluate our model on a synthetic O(5) invariant regression problem $2T_1\to T_0$ in $d=5$ dimensions given by the function $f(x_1,x_2) = \sin (\|x_1\|)-\|x_2\|^3/2 + \frac{x_1^\top x_2}{\|x_1\|\|x_2\|}$. We evaluate EMLP-$\mathrm{SO}(5)$ and EMLP-$\mathrm{O}(5)$ which is also equivariant to reflections. We compare against a standard MLP as well as MLP-Aug that is trained with $\mathrm{O}(5)$ data augmentation. We show the results in \autoref{fig:data_efficiency}.

\textbf{$\mathrm{O}(3)$ Equivariant Task}:
Next we evaluate the networks on the equivariant task of predicting the moment of inertia matrix $\mathcal{I} = \sum_i m_i (x_i^\top x_i I-x_ix_i^\top)$ from $n=5$ point masses and positions. The inputs $X = \{(m_i,x_i)\}_{i=1}^5$ are of type $5T_0+5T_1$ ($5$ scalars and vectors) and outputs are of type $T_2$ (a matrix), both transform under the group. We apply $\mathrm{SO}(3)$ and $\mathrm{O}(3)$ equivariant models to this problem. For the baselines, we implement data augmentation for the standard MLP for this equivariant task by simultaneously transforming the input by a random matrix $R \in \mathrm{O}(3)$ and transforming the output accordingly by the inverse transformation: $\hat{y} = R^\top\mathrm{MLP}(\{(m_i,Rx_i\}_{i=1}^5)R$. This kind of equivariant data augmentation that transforms both the input and the output according to the symmetry is strong baseline.

\begin{figure*}[!ht]
	\centering
	\subfigure{
    \includegraphics[width=.18\linewidth]{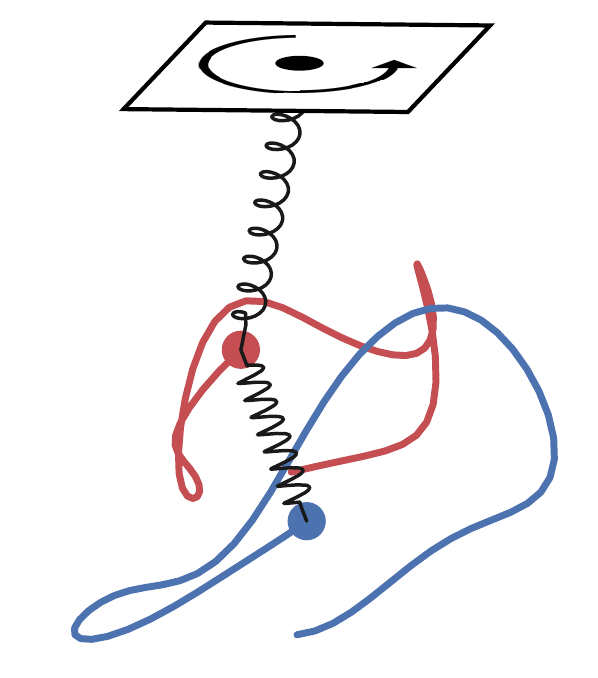}}\hspace{.5cm}
	\subfigure{
    \includegraphics[width=.335\linewidth]{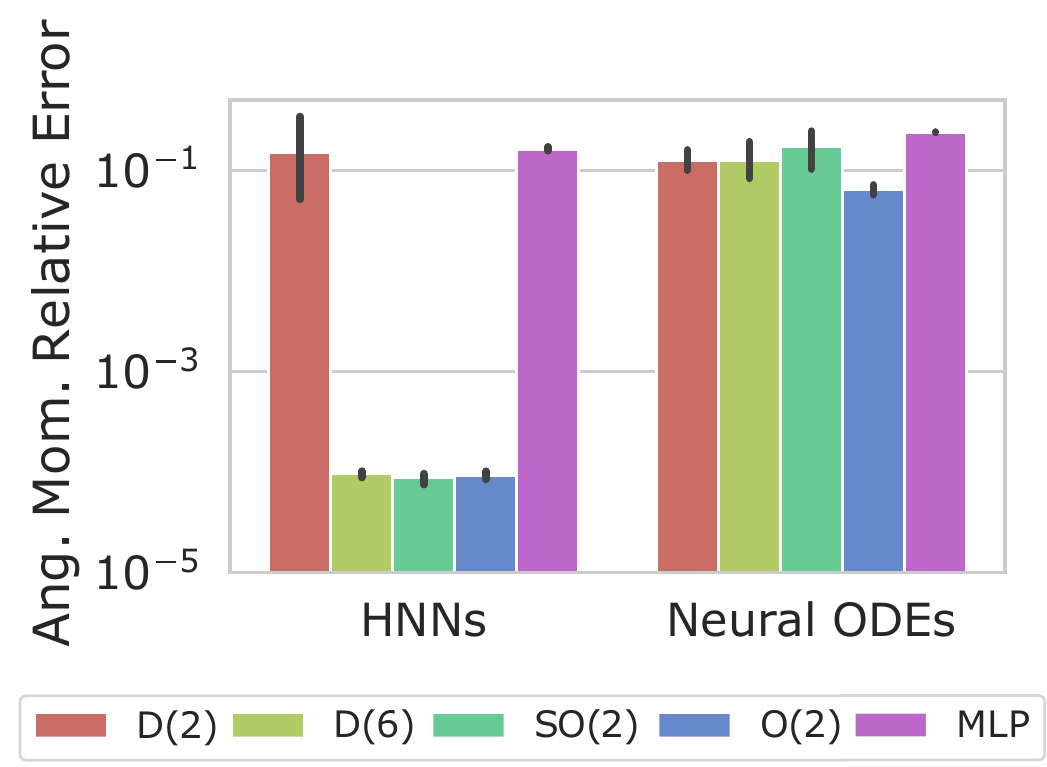}}\hspace{.5cm}
	\subfigure{
    \includegraphics[width=.34\linewidth]{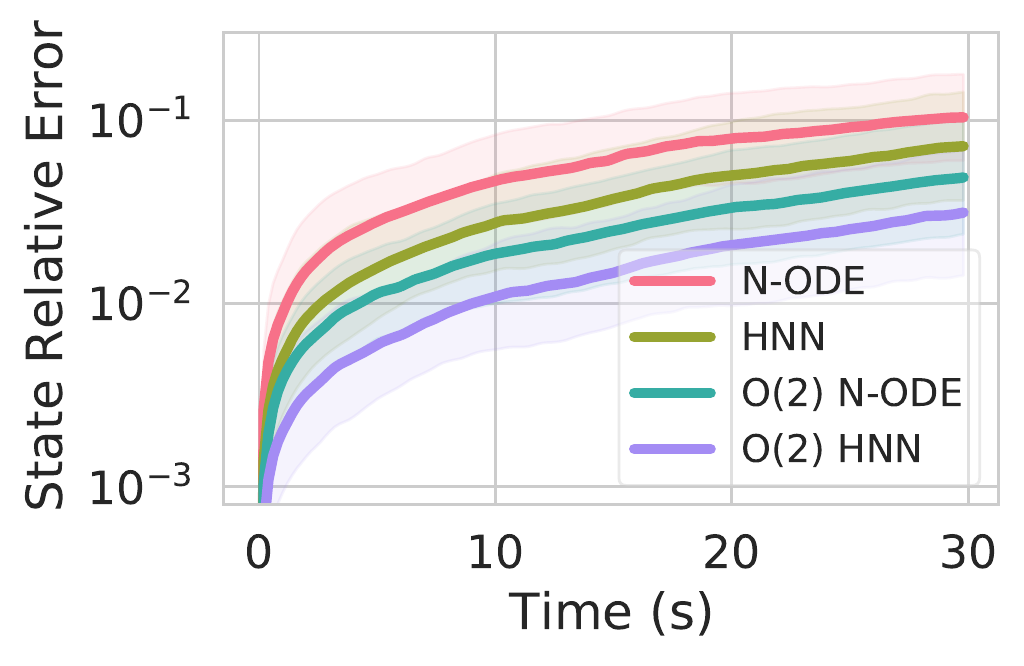}}
	\caption{\textbf{Left}: A double spring pendulum (12s sample trajectory is shown). The system has an $\mathrm{O}(2)$ symmetry about the $z$ axis. \textbf{Middle}: Conservation of angular momentum about the $z$-axis (the geometric mean of the relative error is computed over $30s$ rollouts and averaged across initial conditions). Errorbars are 95\% confidence interval over 3 runs. \textbf{Right}: The relative error in the state as the trajectory is rolled out. Shaded regions show 1 standard deviation in log space across the different trajectories rather than models, showing the variance in the data.
	}
    \label{fig:spring_pendulum}
\end{figure*}

\textbf{Lorentz Equivariant Particle Scattering}:
Testing the ability of the model to handle Lorentz equivariance in tasks relevant to particle physics, we train models
to fit the matrix element in electron muon scattering 
$e^- + \mu^- \to e^- + \mu^-$ which is proportional to the scattering cross-section.
The scattering matrix element is proportional to $|\mathcal{M}|^2\propto$
\begin{equation*}
    [p^{(\mu}\tilde{p}^{\nu)} - (p^\alpha \tilde{p}_\alpha - p^\alpha p_\alpha)\eta^{\mu\nu}][q_{(\mu}\tilde{q}_{\nu)} - (q^\alpha \tilde{q}_\alpha - q^\alpha q_\alpha)\eta_{\mu\nu}]
\end{equation*}

\citep{scattering} where $q_\mu$ and $p_\mu$ are the four momenta for the ingoing electron and muon respectively, while $\tilde{q}_\mu$ and $\tilde{p}_\mu$ are the outgoing momenta, and parentheses $(\mu\nu)$ denotes the symmetrization of indices and repeated indices are contracted. While simple enough express in closed form, the scalar output involves contractions, symmetrization, upper and lower indices, and a metric tensor.
Here the inputs are $4T_{(1,0)}$ and the output is a scalar $T_{(0,0)}$. We evaluate EMLP with equivariance not just to the proper orthochronious Lorentz group $\mathrm{SO}^+(1,3)$ from \citet{bogatskiy2020lorentz}, but also the special Lorentz group $\mathrm{SO}(1,3)$, and the full Lorentz group $\mathrm{O}(1,3)$ and compare a MLP baseline that uses $\mathrm{O}(1,3)$ data augmentation.

As shown in \autoref{fig:data_efficiency}, our EMLP model with the given equivariance consistently outperform sthe baseline MLP trained with and without data augmentation across the different dataset sizes and tasks, often by orders of magnitude.

\subsection{Modeling dynamical systems with symmetries}
Finally we turn to the task of modeling dynamical systems.
For dynamical systems, the equations of motion can be written in terms of the state $\state\in \mathbb{R}^m$ and time $t$ as $d\state/dt = F(\state,t)$. Neural ODEs \citep{chen2018neural} provide a way of learning these dynamics directly from trajectory data. A neural network parametrizes the function $F_\theta$ and the learned dynamics can be rolled out using a differentiable ODE solver $(\hat\state_1, \dots, \hat\state_T) = \mathrm{ODESolve}(\state_0,F_\theta,(t_1,t_2,...,t_T))$
 and fit to trajectory data with the L2 loss 
$L(\theta) = \frac{1}{T} \sum\limits_{t=1}^T ||\hat\state_t -\state_t||_2^2$.

Many physically occurring systems have a Hamiltonian structure, meaning that the state can be split into generalized coordinates and momenta $\state = (\position,\momentum)$, and the dynamics can be written in terms of the gradients of a scalar $\mathcal{H}(z)$ known as the Hamiltonian, which often coincides with the total energy.
$\frac{d\state}{dt} = J \nabla\mathcal{H}$ with 
$J = \begin{bmatrix}0 & I \\-I & 0 \\\end{bmatrix}$.
As shown in \citet{greydanus2019hamiltonian} with Hamiltonian Neural Networks (HNNs), one can exploit this Hamiltonian structure by parametrizing $\hat{\mathcal{H}}_\theta(\state)$ with a neural network, and then taking derivatives to find the implied Hamiltonian dynamics. For problems with Hamiltonian structure HNNs often lead to improved performance, and better energy conservation.

A dynamical system can have \emph{symmetries} such as the symmetries given by $F(\rho(g)\state,t) = \rho(g)F(\state,t)$ for some linear representation,
which is equivariance in the first argument. Meanwhile Hamiltonian dynamics have symmetries according to invariances of the Hamiltonian $\mathcal{H}(\rho(g)\state) = \mathcal{H}(\state)$. Continuous symmetries of the Hamiltonian are of special significance since they produce conservation laws such as conservation of linear and angular momentum or conservation of charge as part of the Noether theorem \citep{noether1971invariant}.

We apply our EMLP model to the task of learning the dynamics of a double pendulum connected by springs in 3D shown in \autoref{fig:spring_pendulum}. The problem exhibits a $\mathrm{O}(2)$ rotational and reflectional symmetry about the $z$-axis as well as Hamiltonian structure. As the state space cannot be traversed by the group elements alone, it is not a homogeneous space, a setting that has been explored very little in the equivariance literature \citep{cohen2018general}.

However, we can readily use EMLP on this problem and we show in \autoref{table:dynamical_systems} and \autoref{fig:spring_pendulum} that exploiting the $\mathrm{O}(2)$ symmetry (and subgroups $\mathrm{SO}(2)$, $\mathrm{D}_6$) with EMLP leads to improved performance for both Neural ODE and HNN models. Furthermore, enforcing the continuous rotation symmetry in the EMLP-HNN models yields conservation of angular momentum about the $z$-axis, a useful property for learned simulations. Interestingly the dihedral group $\mathrm{D}_6$ which is discrete does not satisfy Noether's theorem and yet it still yields approximate angular momentum conservation, but the coarser $\mathrm{D}_2$ symmetry does not. As expected, all Neural ODE models do not conserve angular momentum as Noether's theorem only applies to the Hamiltonians and not to the more general ODEs. While conservation laws from learning invariant Hamiltonians was also explored in \citet{finzi2020generalizing} with LieConv, LieConv models assume permutation equivariance which is broken by the pivot in this system. Because EMLP is general, we can apply it to this non permutation symmetric and non transitively acting rotation group that is embedded in the larger state space.

\begin{table}[t]
    \centering
    
    \begin{tabular}{cccc c}
    &$\mathrm{O}$(2) &$\mathrm{SO}$(2)  &$\mathrm{D}_6$ & MLP\\
     \midrule
     N-ODEs:&\bf{0.019}(1)& 0.051(36)  & 0.036(25) &0.048\\
     HNNs: &\bf{0.012}(2)& 0.015(3)& 0.013(2) & 0.028\\
    \end{tabular}
    \caption{Geometric mean of rollout errors (relative error) over T=$30$s for the various EMLP-$G$ symmetric HNNs and Neural ODEs (N-ODE) vs ordinary MLP HNNs and N-ODEs. Errorbars are 1 standard deviation computed over 3 trials, with notation $.012(2)$ meaning $.012\pm .002$.
    }\label{table:dynamical_systems}
    \vspace{-.4cm}
\end{table}

\section{Discussion}
We presented a  construction for equivariant linear layers that is completely general to the choice of representation and matrix group. Convolutions, deep sets, equivariant graph networks and GCNNs all fall out of the algorithm naturally as solutions for a given group and representation. Through an iterative MVM based approach, we can solve for the equivariant bases of very large representations. Translating these capabilities into practice, we build EMLP and apply the model to problems with symmetry including Lorentz invariant particle scattering and dynamical systems, showing consistently improved generalization.

Though EMLP is not much slower than a standard MLP, dense matrix multiplies in an MLP and our EMLP make it slow to train models the size of convnets or large graph networks which have specialized implementations. With the right techniques, this apparent generality-specialization tradeoff may be overcome. The flexibility of our approach should lower the costs of experimentation and allow researchers to more easily test out novel representations. Additionally we hope that our constraint solver can help launch a variety of new methods for learning symmetries, modeling heterogeneous data, or capturing prior knowledge.

\section*{Acknowledgements} We would like to thank Roberto Bondesan and Robert Young for useful discussions about equivariance and Lie algebra representations. This research is supported by an Amazon Research Award, NSF I-DISRE 193471, NIH R01 DA048764-01A1, NSF IIS-1910266, and NSF 1922658 NRT-HDR: FUTURE Foundations, Translation, and Responsibility for Data Science.

\bibliography{refs}
\bibliographystyle{icml2021}

\clearpage
\appendix
\section{Overview}
In \autoref{sec:app_constraints_proof} we prove that \autoref{eq:constraint_lie} and \autoref{eq:constraint_discrete} are necessary and sufficient conditions to satisfy the symmetry constraint.
In \autoref{sec:krylov_runtime} we prove that the simple iterative MVM method for finding the nullspace converges to the true nullspace with an exponential rate, and we derive the complexity upper bounds for various symmetry groups and representations.
In \autoref{sec:app_bilinear_necessary} we show that using gated nonlinearities or Norm-ReLUs alone are not sufficient for universality. In \autoref{sec:app_bases} we detail the various groups we implement and calculate the equivariant subspaces for different tensor representations.
In \autoref{sec:app_recipe} we detail the steps necessary to extend our implementation to new groups and representations.
\autoref{sec:app_group_products} details some additional rules by which the solutions for group products can be sped up.
Finally, \autoref{sec:app_implementation} and \autoref{sec:app_dataset} describe training hyperparameters and how the datasets were constructed.
\section{Necessary and Sufficient Conditions for Equivariance}\label{sec:app_constraints_proof}

\begin{theorem}
Given a (real) Lie group $G$ with a finite number of connected components, and a representation $\rho$ acting on vector space $V$,
the symmetry constraint 
\begin{equation}
    \forall g\in G: \rho(g)v=v
\end{equation}
for $v\in V$ is satisfied if and only if
\begin{align}
    \forall i=1,..,D:&&d\rho(A_i)v&=0,\label{eq:lie_cons_app}\\
    \forall \ell=1,...,M:&& (\rho(h_\ell)-I)v&=0,\label{eq:dis_cons_app}
\end{align}
where $\{A_i\}_{i=1}^D$ are $D$ basis vectors for the $D$ dimensional Lie Algebra $\mathfrak{g}$ with induced representation $d\rho$, and for some finite collection $\{h_\ell\}_{\ell=1}^M$ of discrete generators.
\end{theorem}

\textbf{Proof:} As shown in \autoref{sec:app_finitely_generated}, elements of a (real) Lie group can be written as $g = \exp{(\sum_i\alpha_iA_i)}\Pi_i h_{k_i}$ for some collection of real valued coefficients $\alpha_i \in \mathbb{R}$ and discrete coefficients $k_i \in [-M,...,M]$ which index the $M$ discrete generators (and their inverses). Note that $M$ is upper bounded $M\le (D+1)+\mathrm{nc}(G)$, by the sum of the dimension and the number of connected components of $G$ and is often much smaller as shown by the examples in \autoref{sec:app_bases}. For subgroups of $S_n$ for example, $M\le n$ \citep{guralnick1989number}. Discrete groups are included as a special case of Lie groups with $D=0$. The forward and backward directions of the proof are shown below.

\textbf{Necessary:}
Assume $\forall g\in G: \rho(g)v=v$. Writing $g=\exp{(\sum_i\alpha_iA_i)}\Pi_i h_{k_i}$, we can freely choose group elements with $k = \varnothing$ to find that
\begin{equation*}
    \forall \alpha: \rho\big(\exp{(\sum_{i=1}^D\alpha_iA_i)}\big)v = \exp{\big(d\rho(\sum_{i=1}^D\alpha_iA_i)\big)}v = v
\end{equation*}
using the correspondence between group and algebra representation \eqref{eq:lie_correspondance}.
From the linearity of $d\rho(\cdot)$, this implies $\exp{\big(\sum_i\alpha_id\rho(A_i)\big)}v = v$. Taking the derivative with respect to $\alpha_i$ at $\alpha=0$, we have that 
\begin{equation}
    \forall i=1,..,D: \quad d\rho(A_i)v=0,
\end{equation}
since the exponential map satisfies $\frac{d}{dt}\exp(tB)|_{t=0} = B$.

Similarly for the discrete constraints we can set $\alpha_i=0$ and $k=[\ell]$ so that $\rho(g)v=\rho(\Pi_i h_{k_i})v = \rho(h_\ell)v=v$. By setting varying $\ell$, we get the $M$ constraints
\begin{equation}
    \forall \ell=1,...,M: \quad (\rho(h_\ell)-I)v=0.
\end{equation}

\textbf{Sufficient:}
Assume \autoref{eq:lie_cons_app} and \autoref{eq:dis_cons_app} both hold. Starting with the continuous constraints, 
exponential map (defined through the Taylor series) satisfies 
\begin{equation*}
    \exp{(B)}v = v+Bv + \frac{1}{2}B^2v +...
\end{equation*}
but if $Bv=0$ then $\exp{(B)}v=v$ since all terms except $v$ are $0$.
Setting $B=\sum_i\alpha_id\rho(A_i) =d\rho(\sum_i\alpha_iA_i)$ which satisfies $Bv=0$, we have that 
\begin{align}\label{eq:cont_sat}
\begin{split}
    \forall \alpha_i: \quad \exp{\big(d\rho(\sum_i\alpha_iA_i)\big)}v &=v\\
     \forall \alpha_i: \quad \rho\big(\exp{(\sum_i\alpha_iA_i)}\big)v&=v.
\end{split}
\end{align}
Similarly for the discrete generators, if $\forall \ell: \rho(h_\ell)v=v$ then
\begin{equation*}
    \forall \ell:\quad v = \rho(h_\ell)^{-1}v = \rho(h_{-\ell})v,
\end{equation*}
and any product satisfies
\begin{equation*}
    \rho(\Pi_{i=1}^N h_{k_i})v = \Pi_{i=1}^N\rho(h_{k_i})v = \big(\Pi_{i=1}^{N-1}\rho(h_{k_i})\big)\rho(h_{k_N})v.
\end{equation*}
since $\rho(h_{k_N})v=v$ we can remove that factor and repeat the argument to get
\begin{equation}\label{eq:disc_sat}
    \rho(\Pi_{i=1}^N h_{k_i})v = \big(\Pi_{i=1}^{N-1}\rho(h_{k_i})\big)v =... = v.
\end{equation}

Putting \autoref{eq:cont_sat} and \autoref{eq:disc_sat} together, we have that for all $\alpha,k$: $\rho(\exp{(\sum_i\alpha_iA_i)}\Pi_i h_{k_i})v=v$. Since every group element can be expressed as $g=\exp{(\sum_i\alpha_iA_i)}\Pi_i h_{k_i}$ for some $\alpha,k$, the equivariance constraint $\forall g\in G: \rho(g)v=v$ is satisfied. 

\section{Krylov Method for Efficiently Finding the Equivariant Subspace}\label{sec:krylov_runtime}
The iterative Krylov subspace algorithm that we use to find the nullspace of the constraint matrix $C$ is a close variant of the iterative methods for finding the largest eigenvectors such as power iteration and Ojas method . 
We need to be able to compute the nullspaces of the massively large constraint matrices $C$ (such as the $(4\times 10^5) \times (7\times 10^4)$ sized matrix for computing the equivariant subspace of $T_7^{S_5}$ in \autoref{sec:app_bases}), making use of efficient structure that allows fast MVMs with $C$.

While the shift and invert strategy for finding small eigenvalues is commonly recommended \citep{ipsen1997computing}, the costs of inversion via conjugate gradients for these massive matrices can make it exceedingly slow in practice. Other methods such as block Lanczos \citep{eberly2004reliable} and Wiedemann \citep{turner2006block} have been explored in the literature for the nullspace problem, but these methods tend to be exceedingly complicated. 

Instead we provide a much simpler algorithm that is also extremely fast: simple gradient descent followed by a small SVD. We prove that gradient descent converges exponentially in this problem, and that with probability $1$ our method converges to the correct nullspace with error $\epsilon$ in $O(\log(1/\epsilon))$ iterations. We then verify this fact empirically. The algorithm is closely related to power iteration \citep{francis1961qr}, Oja's rule \citep{shamir2015stochastic}, and the PCA approaches that are framed as optimization problems \citep{garber2015fast}.

As introduced in \autoref{alg:subspace} we propose the following algorithm for finding the nullspace with rank $r\le r_{\mathrm{max}}$ and then use iterative doubling of $r_{\mathrm{max}}$ until the conditions are met.
\begin{algorithm}[H]
\SetAlgoLined
def \textbf{CappedKrylovNullspace}($C,r_\mathrm{max}$):\\
 $X \sim \mathcal{N}(0,1)^{n\times r_\mathrm{max}}$\\
\While{$L(X) > \epsilon$}{
    $L(X) = \|CX\|^2_F$\\
    $X \gets X - \eta\nabla L$
}
$\tilde{Q},\tilde{\Sigma},\tilde{V} = \mathrm{SVD}(X)$\\
\Return $\tilde{Q}$\\
\caption*{Fast Krylov Nullspace}
\end{algorithm}
Assume then that $r\le r_{\max}$ (abbreviated $r_m$) and that we will use the notation from \autoref{eq:svd} that $C =U\begin{bmatrix} \Sigma & 0 \\ 0 & 0\\\end{bmatrix}\begin{bmatrix}P^\top \\ Q^\top \end{bmatrix}$. The gradients of $L$ are $\nabla_X L = C^\top C X$ and thus gradient descent can be written as the iteration
\begin{equation}
    X_{t+1} = (I-\eta C^\top C)X_t.
\end{equation}
We can write $X$ in terms of the true singular vectors of $C$ (the eigenvectors of $C^\top C$) which form a basis. $Q$ and $P$ are orthogonal ($Q^\top Q=I$, $Q^\top P =0$, $P^\top P=I$) and we define the projections of $X_t$ onto these subspaces, $W_t = Q^TX_t$ and $V_t = P^TX_t$. As orthogonal transformations of isotropic Gaussians are also isotropic Gaussians, the two matrices are initially distributed $W_0 \sim \mathcal{N}(0,1)^{r\times r_m}$  and $V_0 \sim \mathcal{N}(0,1)^{(n-r)\times r_m}$.

\begin{figure}[t]
    \centering
	\includegraphics[width=0.48\textwidth]{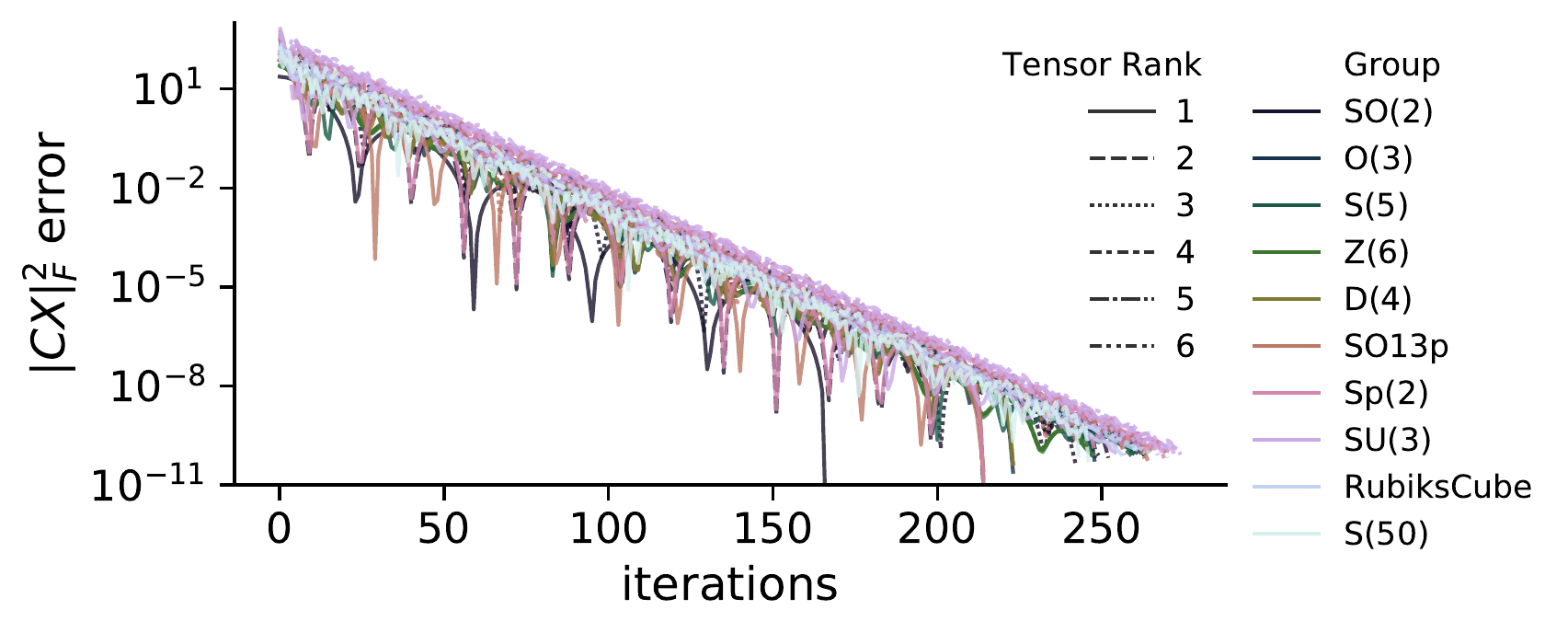}
	\caption{Exponential convergence of \autoref{alg:subspace} shown empirically over a range of groups and tensor representations for $r_\mathrm{max}=20$. In each of these cases, $X$ converges to the limits of floating point precision in $300$ iterations.
	}
    \label{fig:convergence}
\end{figure}

Writing $X_t = QW_t +PV_t$, and noting $C^\top C = P\Sigma^2P^\top$ we can now see the effect of the iteration on the subspaces:
\begin{align*}
    QW_{t+1} + PV_{t+1} &= (I-\eta P\Sigma^2P^\top)\big(QW_t + PV_t)\\
    &= QW_t + P(I-\eta \Sigma^2)V_t.
\end{align*}
Unrolling the iteration, we have that
$W_t = W_0$ and $V_t = P(I-\eta \Sigma^2)^tV_0$. So long as the learning rate is chosen $\eta < 2/\sigma_{\mathrm{max}}^2$ then the iteration will converge exponentially to $X = QW_0$. Given optimal learning rate, the convergence is $T = O(\kappa \log(1/\epsilon))$ where $\kappa = (\frac{\sigma_{\mathrm{max}}}{\sigma_{\mathrm{min}}})^2$. Since $W_0$ is a Gaussian random matrix $\mathcal{N}(0,1)^{r\times r_m}$, it will be full rank $r$ with probability $1$. Therefore, performing a final SVD on $X$ will yield the nullspace $Q$. The runtime of this procedure is $O((M+D)\mathcal{T}r_m\log(\frac{1}{\epsilon})+r_m^2n)$ since each matrix multiply with $C$ and $C^\top$ takes time $(M+D)\mathcal{T}r_m$ where $\mathcal{T}$ is the time for multiplies with the $\rho$ and $d\rho$ matrices with a single vector, and there are $M+D$ such multiplies that go into a single multiply with $C$. Finally the $r_m^2n$ factor is the cost of taking the SVD of $X$ at the end. The exponential convergence is shown empirically across a range of groups in \autoref{fig:convergence}.

If $r>r_\mathrm{max}$ then the SVD output $\tilde{Q}$ is a random projection of rank $r_\mathrm{max}$ of the true nullspace $Q$. Given an unknown $r$, we can simply rerun the algorithm doubling $r_\mathrm{max}$ each time until the rank of $\tilde{Q}$ is less than $r_\mathrm{max}$. Adding up the costs, the total runtime of the algorithm to reach an $\epsilon$ accurate solution for the nullspace is \begin{equation}
    O((M+D)\mathcal{T}r\log(\tfrac{1}{\epsilon})+r^2n).
\end{equation}

To put this runtime into perspective, we can upper bound the runtime to compute the symmetric bases for rank $p$ tensors $T_k$ of any subgroup of the symmetric group $G\le S_m$. Since all $G\le S_m$ can be expressed with $D+M<m$ discrete generators \citep{guralnick1989number}, and axis-wise permutations of the $n=m^k$ sized tensors can be performed in time $\mathcal{T}=m^k$, the runtime is upper bounded by $O(m^kr(m\log(\tfrac{1}{\epsilon}) + r))$.

Similarly for the orthogonal groups $\mathrm{SO}(m)$ and $\mathrm{O}(m)$ with $D=m(m-1)/2$ infinitesmal generators and $M\le 1$ discrete generators, the MVM time can be done in $\mathcal{T}=km^k$ since the infinitesmal generators can be expressed with only $2$ nonzero elements. Putting this together, the symmetric spaces for rank $T_k$ tensors can be solved for in time $O(m^kr(km^2\log(\tfrac{1}{\epsilon}) + r))$, where $r$ is also upper bounded by the Bell numbers $r\le B_k$. Generalizations of the Lorentz group $\mathrm{SO}(p,n-p)$ and $\mathrm{O}(p,n-p)$ have identical runtimes, and similarly for the complex groups $\mathrm{SU}(n)$ and $\mathrm{U}(n)$, as well as the symplectic group $\mathrm{Sp}(n)$.

For the equivariant maps between the popular \emph{irreducible} representations: $\rho = \psi_k \otimes \psi_\ell$ for the group $G=\mathrm{SO}(3)$, $\mathcal{T} = k^2\ell + k\ell^2$ giving the runtime of our method $O((k^2\ell + k\ell^2)r\log(\tfrac{1}{\epsilon})+r^2k\ell)$. Meanwhile for irreducible representations of $G=\mathrm{SO}(2)$ the runtime is a mere $O(r\log(\tfrac{1}{\epsilon})+r^2)$ regardless of $k$ and $\ell$.

\section{Linear Layers and Gated Nonlinearities are Not Universal}\label{sec:app_bilinear_necessary}
Outside of the regular representations where each $\rho(g)$ is a permutation matrix, we cannot necessarily use pointwise nonlinearities. Existing equivariant nonlinearities for this setting such as Norm-ReLU and Gated-Nonlinearity can artificially limit the expressivity of the networks in cases such as when using tensor representations.

\begin{theorem}
Consider equivariant networks built only from equivariant linear layers that map between (direct sums of) tensor representations with features $v\in \bigoplus_{a\in \mathcal{A}} T_a$ (as well as biases) and nonlinearities which act separately on each of the tensors $\sigma: T_a\rightarrow T_a$, or with an additional scalar as $\sigma: T_0 \times T_a\rightarrow T_a$.
There exists groups (like $\mathrm{SO}(2)$ and $\mathrm{O}(3)$) for which these networks cannot approximate even simple equivariant functions.
\end{theorem}

Suppose the base representation $\rho$ of a group $G$ includes elements that satisfy $\exists (g,g'): \rho(g')=-\rho(g)$, which we term the \emph{parity property}. For simplicity assume the representation is orthogonal $\rho = \rho^*$ so that we can talk about rank $k$ tensors rather than rank $(p,q)$ tensors, but the same argument also applies for non-orthogonal representations setting $k=p+q$. Tensor representations $\rho_k(g) = \rho(g)^{\otimes k}$ that have an order $k$ that is \emph{odd} will also have the parity property since $\rho(g')^{\otimes k} = (-1)^k\rho(g)^{\otimes k}$. But because the equivariance constraint holds for all $g\in G$, the following two constraints must also hold for odd $k$:
\begin{align*}
    \rho(g)^{\otimes k}v &= v\\
    -\rho(g)^{\otimes k}v &=v.
\end{align*}
Adding the two constraints together, we have that all equivariant tensors of odd rank (for groups with this property) are $v=0$. This also means that all equivariant linear maps from a tensor with even rank to a tensor with odd rank will be $0$.

This is a property of the equivariance for linear layers for certain groups and is not by itself a problem; however, if the equivariant nonlinearities $\sigma$ act separately on each tensor and preserve its rank $\sigma: T_a\rightarrow T_a$ then all quantities in the network (inputs, outputs, and features) of even order are computationally disconnected from those of odd order. Since the nonlinearities act only on a given tensor and keep its order the same, and linear layers between even and odd are $0$, there can be no nontrivial path between the two.
Similarly for nonlinearities like Gated-Nonlinearities which take in an additional scalar gate as input, so that the nonlinearities are maps $\sigma: T_0\times T_a \rightarrow T_a$ we can extend the result. For these kinds of nonlinearities, features of odd order can depend on inputs and features in previous layers of odd and even rank; however, there is still no path from a feature or input of odd order to a later feature or output of even order.

A simple example is the group $\mathrm{O}(3)$ and the standard vector representation for $R^\top R=I$, $\rho(R)=R$. Suppose the input to the network is the vector $v\in T_1$ and the target to be learned is the scalar $f(v) = \|v\| \in T_0$. Since the input is an odd order tensor and the output is an even order tensor, there can be no nonzero computational path in the network connecting them. Using Norm-ReLU or Gated-Nonlinearities the only valid output of such a network is a constant $c$ which is independent of the input, and the network cannot fit a simple function such as $\|\cdot\|$. 

Of course this limitation extends much beyond this simple example, preventing inner products, matrix vector multiplies, and many other kinds of valid equivariant functions from being expressed, regardless of the size of the network. In fact, using the standard vector representation for all groups $\mathrm{O}(n)$ as well as $\mathrm{SO}(2n)$ satisfies the parity property, and will provably have this limitation. Other groups like the Lorentz group $\mathrm{SO}(1,3)$ and $\mathrm{O}(1,3)$, and the symplectic group $\mathrm{Sp}(n)$ satisfy this property.

\section{Equivariant Bases for Various Groups}\label{sec:app_bases}
In this section we list the dimension of the symmetric bases for various groups and tensor representations that we calculate using our algorithm, and visualize the bases.

\subsection{Discrete Translation Group $\mathbb{Z}_n$}
The discrete translation group (or cyclic group) $\mathbb{Z}_n$ is generated by a single shift permutation $\mathrm{P}[n,1,2,...,n-1]$. Translation equivariant neural networks make use of convolutions which are maps $(T_1 \rightarrow T_1) \cong T_2$ and average pooling $(T_1\rightarrow T_0) \cong T_1$. We recover the $n$ dimensional convolution bases and the $1$ dimensional average pooling element as shown in the table and \autoref{fig:app_bases}. We also show the ranks for higher order tensors, which appear to satisfy $r=n^{k-1}$ for $\mathbb{Z}_n$ and $T_k$. While we are not aware of the higher order equivariant tensor having been derived in the literature, it's not unlikely due to the prominence of the translation group in signal processing.
\begin{table}[H]
\begin{center}

    \begin{tabular}{l|ccccccc}
    
         & $\mathbb{Z}_2$&$\mathbb{Z}_3$&$\mathbb{Z}_4$&$\mathbb{Z}_5$&$\mathbb{Z}_6$&$\mathbb{Z}_7$&$\mathbb{Z}_8$\\
         \cmidrule(lr){1-1}
\cmidrule(lr){2-8}
$T_1$&     1 &    1   &  1   &  1  &   1   &  1   &  1\\
$T_2$&     2  &   3  &   4   &  5  &   6  &   7   &  8\\
$T_3$&     4  &   9   & 16  &  25   & 36   & 49   & 64\\
$T_4$&     8  &  27   & 64  & 125   &216   &343  & 512\\
$T_5$&    16   & 81  & 256  & 625 & 1296 & 2401     & \\
$T_6$&    32 &  243  &1024 & 3125   &      &        & \\
$T_7$&    64  & 729  &4096     &     &        &     &\\
    \end{tabular}
    \caption{Symmetric subspace rank $r$ for tensors $T_k$ of $G=\mathbb{Z}_n$}
\end{center}

\end{table}

\subsection{Permutation Group $\mathbb{S}_n$}
We review the solutions to the permutation group $\mathbb{S}_n$ which were solved for analytically in \citet{maron2018invariant}, which we solve for numerically using our algorithm. As expected, the solutions bases match the limiting size of the $k$th Bell number $B_k$ as $n\rightarrow \infty$. 

However, \citet{maron2018invariant} claim that the size of the basis is always $B_k$ regardless of $n$ and that is not quite correct. The basis derived in \citet{maron2018invariant} is always equivariant, but sometimes it contains linearly dependent solutions, leading to an overcounting when $n$ is small. The fact that the basis cannot always be of size $B_k$ can be seen from the fact that the total dimension of $T_k$ is $n^k$ and the equivariant subspace thus has rank $r\le n^k$. The Bell numbers grow super exponentially in $k$ (about $(k/\log(k))^k$) and therefore given any $n$ they must exceed the maximum $n^k$ for some value of $k$. The place where the original argument of \citet{maron2018invariant} breaks down is when the equivalence classes $\gamma$ may be empty. For example the equivalence class $\gamma_1 = \{\{1\},\{2\},\{3\},\{4\}\}$ corresponds to indices $i_1$,$i_2$,$i_3$,$i_4$ which are all distinct. But for $n<4$ one cannot form a set of four indices that are all distinct, hence $\gamma_1$ is empty for $n<4$. To the best of our knowledge, the dimension of the equivariant subspaces for small $n$ that we report below are the precise values and have not been presented anywhere else.

\begin{table}[H]
\begin{center}
    \begin{tabular}{l|ccccccc|c}
          &$\mathbb{S}_2$&$\mathbb{S}_3$&$\mathbb{S}_4$&$\mathbb{S}_5$&$\mathbb{S}_6$&$\mathbb{S}_7$&$\mathbb{S}_8$&$B_k$\\
         \cmidrule(lr){1-1}
\cmidrule(lr){2-8}\cmidrule(lr){9-9}
          $T_1$&   1  &   1   &  1   &  1  &   1  &   1  &   1&   1\\
$T_2$&   2   &  2   &  2  &   2  &   2  &   2  &  2 &   2\\
$T_3$&   4  &   5   &  5   &  5  &   5  &   5   &  5&  5\\
$T_4$&    8   & 14  &  15 &   15 &   15  &  15  &  15&  15\\
$T_5$&  16  &  41  &  51  &  52  &  52  &  52  &  52 &  52\\
$T_6$&  32 &  122 & 187  & 202  & 203 &  203  &    & 203\\
$T_7$&  64  & 365 &  715 &  855 &  &    &     &       877\\
    \end{tabular}
    \caption{Symmetric Subspace rank $r$ for tensors $T_k$ of $G=\mathbb{S}_n$}
\end{center}
\end{table}
\FloatBarrier

\subsection{Rubik's Cube Group}
Showing the capabilities to apply to unexplored groups and representations, we compute the equivariant bases for linear layers that are equivariant to the action of the Rubik's Cube group. The Rubik's cube group is a subgroup of the permutation group $G<S_{48}$ containing all valid Rubik's cube transformations. The group is extremely large $|G|>4\times 10^{19}$, but is generated by only $6$ generators: $F,B,U,D,L,R$ a quarter turn about the front, back, up, down, left, and right faces. 

We use the standard $48$ dimensional regular representation where each of the $6*(9-1)=48$ facets (the center facets are excluded) is a component. The state of the Rubik's cube is represented by a $48$ dimensional vector where each component is an integer $0-5$ representing the $6$ possible colors each facet can take. Using this representation, the 6 generators can be expressed as permutations and we refer the readers to the code for the (lengthy) values of the permutations. Below we show the dimension of the equivariant basis and the size of the tensors in which the basis is embedded. Note that as the Rubik's cube is a subgroup of $S_{48}$, and has fewer group elements as symmetries, the size of the equivariant basis is larger $2,6,22,...$ vs $1,2,5, ...$. For $T_4$ of size $48^4=5308416$ we were able to run the solver with $r_\mathrm{max}=20$ before running out of GPU memory.
\begin{table}[H]
\begin{center}
    \begin{tabular}{l|cccc}
         &$T_1$& $T_2$&$T_3$&$T_4$\\
         
\cmidrule(lr){2-5}
r& 2 & 6 & 22 & $>$20\\
$48^k$ & 48 &2304 & 110592 & 5308416 \\
    \end{tabular}
    \caption{Symmetric Subspace rank $r$ for tensors $T_k$ of Rubik's Cube Group}
\end{center}
\end{table}
\FloatBarrier

\subsection{Continuous Rotation Groups $\mathrm{SO}(n)$ and $\mathrm{O}(n)$}
The special orthogonal group $\mathrm{SO}(n)$ and the orthogonal group $\mathrm{O}(n)$ are continuous Lie groups have the Lie algebra
\begin{equation*}
    \mathfrak{o}(n) = \mathfrak{so}(n) = T_\mathrm{id}\mathrm{SO}(n) = \{A\in \mathbb{R}^{n\times n}: A^\top=-A \}
\end{equation*}
of dimension $D=n(n-1)/2$. The orthogonal group can be constructed with the additional discrete generator $h = \begin{bmatrix}-1 & 0 \\ 0 & I_{n-1}\\
\end{bmatrix}$ that has $\mathrm{det}(h)=-1$.
\begin{table}[H]
\begin{center}
    \begin{tabular}{l|cccccc}
         & SO(2)&SO(3)&SO(4)&SO(5)&SO(6)&SO(7)\\
         \cmidrule(lr){1-1}
\cmidrule(lr){2-7}
 $T_2$&     2  &   1  &   1  &   1   &  1   &  1\\
 $T_3$&     0  &   1  &   0  &   0   &  0  &   0\\
 $T_4$&     6  &   3 &    4  &   3  &   3  &   3\\
 $T_5$&     0  &   6  &   0  &   1  &   0 &    0\\
 $T_6$&    20  &  15  &  25  &  15  &  16  &   15 \\
 $T_7$&     0 &   36  &   0  &   15   &      &  \\
 $T_8$&    70  &  91  &   196    &     &      &    \\
    \end{tabular}
    \caption{Symmetric subspace rank $r$ for tensors $T_k$ of $G=\mathrm{SO}(n)$}
\end{center}
\end{table}
\FloatBarrier
We omit the first row $T_1$ since all the values are $0$.
The additional basis element along the diagonal $n=k$ can be recognized as the well known anti-symmetric Levi-Civita symbol $\varepsilon_{ijk\ell...}$.
However this basis element does not respect orientation reversing isometries, and is thus absent in the equivariant basis for $\mathrm{O}(n)$:
\begin{table}[H]
\begin{center}
    \begin{tabular}{l|ccccccc}
         & O(2)&O(3)&O(4)&O(5)&O(6)&O(7)\\
         \cmidrule(lr){1-1}
\cmidrule(lr){2-7}
 $T_2$&     1  &   1  &   1  &   1   &  1   &  1   \\
 $T_3$&     0  &   0  &   0  &   0   &  0  &   0  \\
 $T_4$&     3  &   3 &    3  &   3  &   3  &   3  \\
 $T_5$&     0  &   0  &   0  &   0  &   0 &    0 \\
 $T_6$&    10  &  15  &  15  &  15  &  15  &  15  \\
 $T_7$&     0 &   0  &   0  &   0   &      &      \\
 $T_8$&    35  &  91  & 105      &     &      &      \\
    \end{tabular}
    \caption{Symmetric subspace rank $r$ for tensors $T_k$ of $G=\mathrm{O}(n)$}
\end{center}
\end{table}
\FloatBarrier

\subsection{Lorentz Groups $\mathrm{SO}^+(1,3)$, $\mathrm{SO}(1,3)$, $\mathrm{O}(1,3)$}
The Lorentz group is defined as the set of matrices that preserve the Lorentz metric $\eta$: $\mathrm{O}(1,3) = \{L\in \mathbb{R}^{4\times 4}: L^\top \eta L = \eta\}$. Differentiating, one gets the $D=6$ dimensional Lie algebra $\mathfrak{so}(1,3) = \{A\in \mathbb{R}^{4\times 4}: A^\top \eta +\eta A=0\}$. The full Lorentz group $\mathrm{O}(1,3)$ has four connected components. 

The identity component of the Lorentz group $\mathrm{SO}^+(1,3)$ is just the exponential of the Lie algebra $\mathrm{SO}^+(1,3) = \exp (\mathfrak{o}(1,3))$. The subgroup $\mathrm{SO}(1,3)$ of $\mathrm{O}(1,3)$ with determinant $1$ can be constructed with the additional generator $h_1 = -I$ (which combines time reversal with a parity transformation), and the full Lorentz group $\mathrm{O}(1,3)$ includes $h_1$ as well as the generator $h_2 = \begin{bmatrix}-1 & 0 \\ 0 & I_{3}\\
\end{bmatrix}$ that reverses time only. As these groups are not orthogonal, we must distinguish $T_{(a,b)}$ from $T_{(a+b,0)}$. Below we show the number of basis vectors for $T_{(k,0)}$ which for these $3$ groups is the same number as for $T_{(k-i,i)}$ although the bases elements are distinct.

\begin{table}[H]
\begin{center}
\setlength\tabcolsep{2pt}
    \begin{tabular}{l|cccccccc}
         & $T_{(2,0)}$&$T_{(3,0)}$&$T_{(4,0)}$&$T_{(5,0)}$&$T_{(6,0)}$&$T_{(7,0)}$&$T_{(8,0)}$\\
         \cmidrule(lr){1-1}
\cmidrule(lr){2-8}
 $\mathrm{SO}^+(1,3)$&   1  &   0  &   4   &  0   &  25  &0 &196 \\
 $\mathrm{SO}(1,3)$&     1  &   0  &   4   &  0  &   25 &0&196 \\
 $\mathrm{O}(1,3)$&    1 &    0  &   3  &   0  &   15 &0&105 \\
    \end{tabular}
    \caption{Symmetric subspace rank $r$ for tensors $T_{(k,0)}$ for the Lorentz groups.}
\end{center}
\end{table}

\subsection{Symplectic Group $\mathrm{Sp}(n)$}
Similar to the orthogonal group and the Lorentz group, the symplectic group is defined through the perservation of a quadratic form. 
\begin{equation*}
    \mathrm{Sp}(n) = \{M\in \mathbb{R}^{2n\times 2n}: M^\top \Omega M = \Omega\},
\end{equation*}
 where $\Omega = \begin{bmatrix}0 & I_n \\ -I_n & 0\\
\end{bmatrix}$ and is often relevant in the context of Hamiltonian mechanics and classical physics. The quadratic form $\Omega$ can be interpreted as a measurement of oriented area (in phase space) and is preserved by the evolution of many systems. The $D=n(2n+1)$ dimensional Lie algebra satisfies
\begin{equation*}
    \mathfrak{sp}(n) = \{A\in \mathbb{R}^{2n\times 2n}: A^\top \Omega +\Omega A = 0\},
\end{equation*}
and any element in $\mathrm{Sp}(n)$ can be written $M = \mathrm{exp}(A_1)\mathrm{exp}(A_2)$ for some $A_1,A_2 \in \mathfrak{sp}(n)$.
\begin{table}[H]
\begin{center}
    \begin{tabular}{l|ccccccc}
         & Sp(1)&Sp(2)&Sp(3)&Sp(4)&Sp(5)&Sp(6)\\
         \cmidrule(lr){1-1}
\cmidrule(lr){2-7}
 $T_{(2,0)}$&     1  &   1  &   1  &   1   &  1   &  1   \\
 $T_{(3,0)}$&     0  &   0  &   0  &   0   &  0  &   0  \\
 $T_{(4,0)}$&     2  &   3 &    3  &   3  &   3  &   3  \\
 $T_{(5,0)}$&     0  &   0  &   0  &   0  &   0 &    0 \\
 $T_{(6,0)}$&    5  &  14  &  15  &  15  &    &   \\
 $T_{(7,0)}$&     0 &   0  &   0  &      &      &      \\
 $T_{(8,0)}$&    14  &  84  &      &     &      &      \\
    \end{tabular}
    \caption{Symmetric subspace rank $r$ for tensors $T_{(k,0)}$ of $G=\mathrm{Sp}(n)$}
\end{center}
\end{table}
Although for large values of $n$, the dimension of the basis for $\mathrm{Sp}(n)$ becomes similar to that of its subgroup $\mathrm{O}(n)$, the basis elements themselves are quite different. Like the Lorentz groups, different ways of distributing the rank between the base vector space and its dual as $T_{(k-i,i)}$ for different $i$ yields different solutions for the equivariant basis.
\FloatBarrier
\subsection{Special Unitary Group $\mathrm{SU}(n)$}
Using a complex valued SVD and replacing the objective in the iterative algorithm $L(Q) = \|CQ\|_F^2=\mathrm{Tr}(Q^\top C^\top C Q)$ with $L(Q) = \mathrm{Tr}(Q^\dagger C^\dagger C Q)$ where $\dagger$ is the complex conjugate transpose, we can apply our method to solve for the equivariant bases for complex groups such as the special unitary group $\mathrm{SU}(n)$ relevant for the symmetries of the standard model of particle physics. 

The group can be defined 
\begin{equation*}
    \mathrm{SU}(n)=\{U\in \mathbb{C}^{n\times n}: U^\dagger U=1, \ \ \mathrm{det}(U)=1\}.
\end{equation*} The Lie algebra of dimension $D=n^2-1$ satisfies
\begin{equation*}
    \mathfrak{su}(n) = \{A\in \mathbb{C}^{n\times n}: A^\dagger =-A,  \ \ \mathrm{Tr}(A)=0\}.
\end{equation*}
The group is contained in the image of the exponential map, $\mathrm{exp}(\mathfrak{su}(n)) = \mathrm{SU}(n)$. Since the size of the basis differs between $T_{(4,1)}$ and $T_{(3,2)}$ for example, we show the solutions for only a selection of tensor ranks. It may also be useful to consider \emph{anti}-linear maps, but we leave this to future work.

\begin{table}[H]
\begin{center}
\setlength\tabcolsep{2pt}
    \begin{tabular}{l|cccccccc}
         & $T_{(3,0)}$&$T_{(3,1)}$&$T_{(3,2)}$&$T_{(3,3)}$&$T_{(4,0)}$&$T_{(4,1)}$& $T_{(4,2)}$& \\
         \cmidrule(lr){1-1}
\cmidrule(lr){2-8}
 SU(2)& 0&2&0&5&2&0&5\\
 SU(3)& 1&0&0&6&0&3&0  \\ 
 SU(4)& 0&0&0&6&1&0&0    \\
 
    \end{tabular}
    \caption{Symmetric subspace rank $r$ for tensors $T_{(q,p)}$ of $G=\mathrm{SU}(n)$}
\end{center}
\end{table}
\FloatBarrier
\section{Recipe for Use}\label{sec:app_recipe}
Below we outline the minimum required steps for adding new groups and representations to our existing implementation written in Jax \citep{jax2018github}.

\textbf{Adding new groups}:
\begin{enumerate}
    \item Specify a sufficient set of $M$ discrete generators and their base representation as a matrix $\rho(h_i)$ or as a matrix vector multiply $v\rightarrow \rho(h_i)v$ for each $i=1,2,...,M$
    \item Specify a basis for the Lie algebra (if any) and its base representation as a matrix $d\rho(A_i)$ or as a matrix vector multiply $v\rightarrow d\rho(A_i)v$ for each $i=1,2,...,D$
\end{enumerate}
We walk through these steps and provide examples in 
\textbf{Adding new representations $\tilde{\rho}$ to existing groups}:
\begin{enumerate}
    \item Specify $\tilde{\rho}(h_i)$ as a function of $\rho(h_i)$
    \item Define $\mathrm{dim}(V)$, $==$, and $\mathrm{hash}$ functions for the representation
\end{enumerate}

We provide more detailed instructions along with examples for implementing new groups at \url{https://emlp.readthedocs.io/en/latest/notebooks/3new_groups.html} and new representations at \url{https://emlp.readthedocs.io/en/latest/notebooks/4new_representations.html}.

Given that a base representation $\rho$ is \textit{faithful}, all representations can be constructed as functions of this $\rho$ whatever it may be.

A specification of $\tilde{\rho}(h) = f(\rho(h))$ induces the Lie algebra representation $d\tilde{\rho}(A)$ which may be computed automatically using autograd Jacobian vector products (JVP), without needing to specify it manually. $d\tilde{\rho}(A) = \mathrm{JVP}(f,I,d\rho(A))$ which corresponds in math terms to the operation  $d\tilde{\rho}(A) = Df|_{\rho(h)=I}(d\rho(A))$.

\section{Group Products}\label{sec:app_group_products}
Many of the relevant groups for larger problems like 2D arrays, GCNNs, point clouds \citep{fuchs2020se}, sets of images \citep{maron2020learning}, and hierarchical structures \citep{wang2020equivariant} have multiple distinct group substructures. 2D translation symmetry is the group $G=\mathbb{Z}_n\times \mathbb{Z}_n = \mathbb{Z}_n^2$, GCNNs have $G=H\ltimes \mathbb{Z}_n^2$ point clouds typically have $G=S_n \times \mathrm{E}(3)$, sets of images have the symmetry $S_m \times \mathbb{Z}_n^2$, and a voxelized point cloud network could be $(S_n \times \mathrm{E}(3)) \wr \mathbb{Z}_m^3$. Here these symbols for combining groups are the direct product $(\times)$, semi-direct product $(\ltimes)$ and wreath product $(\wr)$. An additional asymptotic speedup can be achieved for groups that are constructed using these structures by exploiting knowledge about how solutions for the larger group depend on the solutions for the constituent groups. 

Suppose representation $\rho_a$ of the group $G_a$ acts on the space $V_a$ which has the symmetric basis $Q_a$, and $\rho_b$ of $G_b$ acts on $V_b$ with the symmetric basis $Q_b$. 

\textbf{($\bm{\times}$):} As shown in \citep{maron2020learning}, the equivariant basis for $G_a\times G_b$ with rep $\rho_a \otimes \rho_b$ can be written as the Kronecker product $Q_{ab} = Q_a\otimes Q_b$.

\textbf{($\bm{\wr}$):} In \citep{wang2020equivariant} it was worked out that the equivariant basis for $G_a \wr G_b$ with rep $(\rho_a \wr \rho_b) \otimes (\rho_a \wr \rho_b)^*$ satisfies $\text{unvec}(Q_{ab}[\alpha, \beta]) =  \text{unvec}(Q_a \beta) \otimes I +\mathds{1}\mathds{1}^\top \otimes \text{unvec}(Q_b \alpha)$ where $\alpha,\beta$ are coefficient vectors of size $r_a$,$r_b$ and the $\text{unvec}(\cdot)$ operation reshapes a vector into a matrix.

\section{Parametrizing Lie Groups with Continuous and Discrete Generators}\label{sec:app_finitely_generated}
According to \citet{winkelmann2003dense}, every real connected Lie group $G$ of dimension $D$ contains a dense subgroup $H\le G$ generated by $D+1$ elements (which can be constructed explicitly by sampling elements in a neighborhood of the identity).
Since $H$ is dense in $G$, for every element $g\in G$ has a neighborhood $\mathcal{N}(g)$ which contains an element $h\in H$. Since $h\in \mathcal{N}(g)$ it must also be true that $gh^{-1} \in \mathcal{N}(\mathrm{id})$ is in a neighborhood of the identity. Since the exponential map $\exp$ is a bijection for a neighborhood around the identity, it must be possible to express $gh^{-1}=\exp{(A)}=\exp{(\sum_i\alpha_iA_i)}$ for some $A\in \mathfrak{g}$ that we can then write in terms of the basis elements $A_i$. Rearranging, and expressing $h$ in terms of the finite generators for $H$, $h=\Pi_i h_{k_i}$, we have that any $g\in G$ can be written
\begin{equation}
    g = \exp{(\sum_i\alpha_iA_i)}\Pi_i h_{k_i}.
\end{equation}
or more succinctly $G=\exp{(\mathfrak{g})}H$. It is likely that the result can also be extended to other fields like $\mathbb{C}$ such as by embedding the group in a higher dimensional real group, but we focus on the real case here. For groups with multiple connected components, we can apply the same result with at most one additional discrete generator for each of the connected components.
\section{Implementation Details}\label{sec:app_implementation}

While there is substantial freedom in the chosen representation for a given feature layer of a neural network, for maximum expressiveness in the subsequent neural network architecture given a fixed channel budget,
we suggest allocating the channels to the different representations in each layer with a uniform allocation heuristic. Progressing from lower dimensional representations to higher dimensional ones, the multiplicity of the representation should be chosen so that the number of channels for associated with each is approximately the same. We use this heuristic for the equivariant networks of each experiment in the paper.

For the three synthetic experiments, we use networks constructed with $3$ EMLP layers each with $c=384$ channels, followed by a single equivariant linear layer mapping to the output type. The baseline MLPs also have $3$ hidden layers of size $c=384$ each. We train all models with batchsize $500$, and learning rate $3\times 10^{-3}$ with the Adam optimizer \citep{kingma2014adam}. We train for a total of $\min(900000/N,1000)$ epochs where $N$ is the size of the dataset, which we found was ample for convergence of both models in all cases. The training time for the EMLP is a couple minutes, while the MLP model trains in $<1$ minute.

For the modeling of the double spring pendulum dynamical system we use the same hyperparameters except with $c=128$ for all models. For the numerical integrator, we use the adaptive RK integrator that is default to Jax with tolerance $2\times 10^{-6}$. We measure relative error as $\mathrm{relative\_error}(a,b) = \|a-b\|/(\|a\|+\|b\|)$. For calculating state relative error, the state is the full vector $z$ of both position and momentum. We train for a total of $2000$ epochs, long enough for each of the models to converge.  \section{Datasets}\label{sec:app_dataset}
We generate the $\mathrm{O}(5)$ invariant dataset using the function $f(x_1,x_2) = \sin (\|x_1\|)-\|x_2\|^3/2 + \frac{x_1^\top x_2}{\|x_1\|\|x_2\|}$ and sampling $x_1,x_2 \sim \mathcal{N}(0,1)^5$.
We generate the $\mathrm{O}(3)$ equivariant inertia dataset by computing $\mathcal{I} = \sum_{i=1}^5 m_i (x_i^\top x_i I-x_ix_i^\top)$ for $x_i\sim \mathcal{N}(0,1)^3$ and sampling positive masses by passing random entries through a softplus: $m_i \sim \mathrm{Softplus}(\mathcal{N}(0,1))$.

For the Lorentz invariant particle interaction dataset we calculate the targets $y=4[p^{(\mu}\tilde{p}^{\nu)} - (p^\alpha \tilde{p}_\alpha - p^\alpha p_\alpha)\eta^{\mu\nu}][q_{(\mu}\tilde{q}_{\nu)} - (q^\alpha \tilde{q}_\alpha - q^\alpha q_\alpha)\eta_{\mu\nu}]$ from the sampled momenta $p_\mu,\tilde{p}_\mu,q_\mu,\tilde{q}_\mu \sim \mathcal{N}(0,1/4^2)$.

For each of these datasets we separate out a test set of size $5000$ and a validation set of size $1000$ which we use for early stopping.

For the double spring dynamical system, we generate the ground truth trajectories using the Hamiltonian dynamics of the Hamiltonian 
\begin{equation*}
    H(x_1,x_2,p_1,p_2) = V(x_1,x_2)+T(p_1,p_2)
\end{equation*}
where $T(p_1,p_2) = \|p_1\|^2/2m_1 + \|p_2\|^2/2m_2$ and $V(x_1,x_2)=$
\begin{equation*}
    \tfrac{1}{2}k_1(\|x_1\|-\ell_1)^2 + \tfrac{1}{2}k_2(\|x_1-x_2\|-\ell_2)^2 + m_1g^\top x_1 + m_2g^\top x_2.
\end{equation*}

The constants are chosen $m_1=m_2=k_1=k_2=\ell_1=\ell_2=1$. The gravity direction is down $g = [0,0,1]$. We sample the initial conditions from the distribution
$x_1 \sim [0,0,-1.5] +\mathcal{N}(0,.2^2)^3$, $x_2\sim [0,0,-3] +\mathcal{N}(0,.2^2)^3$ and $p_1,p_2 \sim \mathcal{N}(0,.4^2)^3$.
We integrate these systems for a time $T=30s$ and for each initial condition we select a randomly chosen $1s$ chunk (evaluated at five $0.2s$ intervals) as the training data. We generate $1500$ trajectory chunks which we split up into $500$ for each of the train, validation, and test sets.

\end{document}